\documentclass[preprint,12pt]{elsarticle}

\usepackage{amsmath,amsthm,amsfonts,amssymb}

\usepackage{hyperref}
\newcommand{\vs}[1]{\textcolor{blue}{\textbf{(vs:)} #1}}
\usepackage{colortbl}
\usepackage{dialogue}
\usepackage[]{mdframed}
\usepackage[most]{tcolorbox}
\usepackage[normalem]{ulem}
\usepackage{tkz-kiviat,numprint,fullpage} 
\usetikzlibrary{arrows}
\usepackage{pgfplotstable}
\pgfplotsset{compat=1.18}

\definecolor{lightgrey}{rgb}{0.925, 0.925, 0.925}
\newcommand{\codebox}[1]{\texttt{\colorbox{lightgrey}{#1}}}

\newcommand\ColorBox[1]{\textcolor{#1}{\rule{2ex}{2ex}}}

\usepackage{cleveref}

\crefname{mytype}{Test Box}{Test Boxes}
\newtcolorbox[use counter=mytype]{testbox}[2]{
  enhanced,
  label type=mytype,
  label = #1,
  title = Test~#1:~#2
}

\newtcolorbox{myboxVS}[2][]{
  enhanced,
  attach boxed title to top left={yshift=-0.3\baselineskip-0.4pt,xshift=2mm},
  colframe=black,
  colback=white,
  fonttitle=\bfseries,
  colbacktitle=white,
  coltitle=black,
  boxed title style={
    boxrule=0pt,
    colframe=white,
    },
  title=#2,
  #1, breakable}

\newtcolorbox{myframe}[2][]{%
  enhanced,colback=white,colframe=black,coltitle=black,
  sharp corners,boxrule=0.4pt,
  fonttitle=\itshape,
  attach boxed title to top left={yshift=-0.3\baselineskip-0.2pt,xshift=3mm},
  boxed title style={tile,size=minimal,left=0.5mm,right=0.5mm,
    colback=white,before upper=\strut},
  title=#2,#1
}

\newcommand{\RNum}[1]{\uppercase\expandafter{\romannumeral #1\relax}}

\journal{Energy and AI}

\begin{document}

\begin{frontmatter}



\title{A Reliable Knowledge Processing Framework for Combustion Science using Foundation Models}


\author{Vansh Sharma and Venkat Raman}

\affiliation{organization={Department of Aerospace Engineering, University of Michigan},
            city={Ann Arbor},
            postcode={48109-2102}, 
            state={MI},
            country={USA}}

\begin{abstract}
This research explores the integration of large language models (LLMs) into scientific data assimilation, focusing on combustion science as a case study. Leveraging foundational models integrated with Retrieval-Augmented Generation (RAG) framework, the study introduces an approach to process diverse combustion research data, spanning experimental studies, simulations, and literature. The multifaceted nature of combustion research emphasizes the critical role of knowledge processing in navigating and extracting valuable information from a vast and diverse pool of sources.
The developed approach minimizes computational and economic expenses while optimizing data privacy and accuracy. It incorporates prompt engineering and offline open-source LLMs, offering user autonomy in selecting base models. The study provides a thorough examination of text segmentation strategies, conducts comparative studies between LLMs, and explores various optimized prompts to demonstrate the effectiveness of the framework. By incorporating an external vector database, the framework outperforms a conventional LLM in generating accurate responses and constructing robust arguments. Additionally, the study delves into the investigation of optimized prompt templates for the purpose of efficient extraction of scientific literature.
The research addresses concerns related to hallucinations and false research articles by introducing a custom workflow developed with a detection algorithm to filter out inaccuracies. Despite identified areas for improvement, the framework consistently delivers accurate domain-specific responses with minimal human oversight. The prompt-agnostic approach introduced holds promise for future deliberations.
The study underscores the significance of integrating LLMs and knowledge processing techniques in scientific research, providing a foundation for advancements in data assimilation and utilization.

\end{abstract}

\begin{graphicalabstract}

\centering
\includegraphics[width=1.0\textwidth]{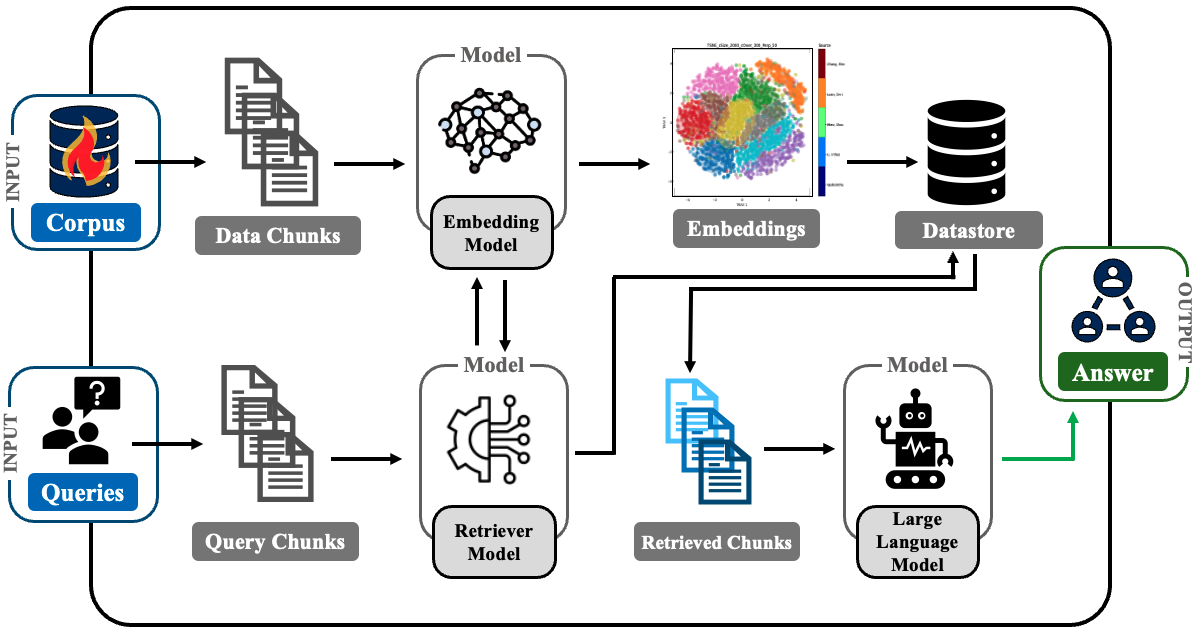}
\end{graphicalabstract}

\begin{highlights}
\item A knowledge processing framework using the retrieval-augmented generation approach is developed for a given domain using archival articles.

\item Comparative study indicates that integrating the framework with a large language model notably enhances accuracy and the quality of arguments.

\item Mitigation strategies for hallucination and the generation of false research articles are proposed and analyzed.

\item Task-specific data containerization is crucial for generating responses with accurate citations and increased precision in arguments. 

\end{highlights}


\end{frontmatter}


\section{Introduction}
With the recent emergence of large language models (LLM), generative artificial intelligence (AI) has provided a new path for scientific data assimilation and use. In particular, so-called scientific foundational models \cite{aurora_gpt} have the potential to provide adaptive ML tools for scientific field-specific research. The purpose of this work is to demonstrate a component of such foundation models, which is the incorporation of existing research literature into refine LLMs, for the specific application of combustion science. The field of combustion is multi-faceted, involving a vast research enterprise that generates data, information and knowledge from a wide-variety of sources including: a) experimental data from lab-, pilot- and device-scale studies, b) simulations that span the range of length and time scales, and c) extensive condensed information in the form of models as well as articles (journals, theses, reports, etc.)  
Given the diverse origins and formats of the information in combustion research, effective knowledge processing becomes paramount. The need arises to sift through, analyze, and synthesize data from experimental studies, simulations, models, and an extensive body of literature. Knowledge processing techniques, including advanced algorithms, become indispensable tools to extract meaningful patterns, correlations, and trends from this large and varied pool of information. This not only facilitates a deeper understanding of foundational physics but also enables researchers to uncover novel insights and optimize combustion processes for various applications. In essence, the multifaceted nature of combustion research underscores the critical role of knowledge processing in navigating and extracting valuable information from the vast and diverse sources within the field.

Earlier language models, such as Eliza (1966)~\cite{eliza} can generate satisfactory responses to only a certain type of prompts. Today, internet is buzzing with chat-agents similar to ChatGPT~\cite{chatGPT}, that are powered by LLMs, such as GPT-4~\cite{openai2023gpt4} and capable of generating exceptional responses to wider variety of prompts. Utilizing the transformer based architecture~\cite{attnIsAllYouNeed}, these models are able to comprehend evolution of information in the form of words, their context with different sentence structures and emotions based on usage. Enabling this learning requires LLMs to be exposed to large corpus of extensive, diverse, and unlabeled data in the form of \emph{embeddings}. An embedding involves transforming a discrete, categorical variable into a vector composed of continuous numerical values~\cite{Mikolov2013EfficientEO}. When applied within the realm of natural language processing (NLP), embeddings refer to compact vectorized representations of words and sentences, typically with low dimensions. Embedding enables dimensionality reduction and enhanced clustering of data categories in the transformed space. LLMs trained on such large databases are referred to as {\emph{foundational models}}. To develop task-specific LLMs, foundation models with pre-trained weights are exposed to different strategies. For instance, by \textbf{fine-tuning with LoRA}~\cite{hu2021lora}, which is typically employed in situations where the task or domain is well-defined and clear, and there is an abundant availability of annotated data for training. Another effective approach is \textbf{reinforcement learning from human feedback (RLHF)}~\cite{ziegler2020finetuning_RLHF}, suited for tasks requiring precision and diverse input sets. RLHF excels in scenarios with limited datasets, allowing the model to generalize effectively through learning from a broad range of inputs via human feedback.

These techniques are direct learning mechanisms for LLMs that use task- or instruction-specific dataset to tune the model weights in the last layer or human-labeled dataset with a continuous learning reward model. Developing such curated data involves significant cost. An alternative, less data-intensive approach is prompt engineering, where carefully designed prompts or inputs in the form of text (instructions or questions) or even images are used to generate high-quality outputs from LLMs. \textbf{Prompt engineering} is ideal for tasks demanding a considerable degree of accuracy and clearly-defined results. In certain scenarios, prompt engineering proves instrumental in enhancing the fine-tuned model's performance, offering additional guidance to the model during the inference process. A meticulously crafted prompt greatly enhances model performance and efficiency by offering clear instructions, reducing ambiguity, and directing the model's understanding \cite{gao-2023}. One should treat prompts as a programming mechanism, as they directly influence output generation and interactions with language models. In the context of natural language processing, these inputs may specify constraints such as tone, style, or response length when necessary.
Prompt engineering involves different strategies, such as zero-shot prompts, few-shot prompts or chain-of-thought prompts, depending on the model behavior. Tuning prompts is critical, and small models can be used to encode the text prompt and generate task-specific virtual tokens~\cite{liu2022ptuning}.

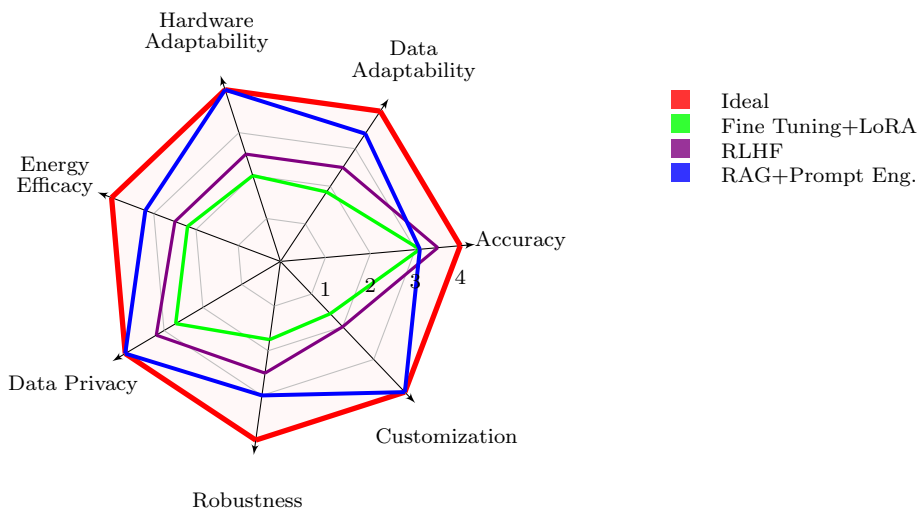
\begin{figure}[hbt!]
 \centering
\begin{tikzpicture}[label distance=.35cm, rotate=05, scale=0.40, font=\fontsize{8}{8}\selectfont]                    
\tkzKiviatDiagram[radial=5, lattice=4, gap=1.5, step=1, label space=7.8cm]%
    {Accuracy,
    Data \\ Adaptability,
    Hardware Adaptability,
    Energy Efficacy,
    Data Privacy,
    Robustness,
    Customization}
\tkzKiviatLine[line width=2pt, color=red, fill=red!05,label=idealA]%
                (4, 4, 4, 4, 4, 4, 4); label{p4}
\tkzKiviatLine[line width=1.3pt,color=green,fill=]%
                (3.1, 1.85, 2, 2.2, 2.7, 1.75, 1.6); label{p5}
\tkzKiviatLine[line width=1.2pt,color=violet,fill=]%
                (3.5, 2.5, 2.5, 2.5, 3.2, 2.5, 2); label{p6}
\tkzKiviatLine[line width=1.5pt,color=blue, fill=]%
                (3.1, 3.4, 4, 3.2, 4, 3, 4); label{p7}
\tkzKiviatGrad[prefix = , unity=1, suffix=](0)
    
\node[anchor=south west,xshift=20pt,yshift=120pt] at (current bounding box.south east) 
{
\begin{tabular}{@{}lp{4cm}@{}}
\ColorBox{red!80} & Ideal \\
\ColorBox{green!80} & Fine Tuning+LoRA \\
\ColorBox{violet!80} & RLHF \\
\ColorBox{blue!80} & RAG+Prompt Eng. \\
\end{tabular}
};
\end{tikzpicture}
\caption{Radar diagram comparing different LLM optimization strategies for information extraction task. The current work focuses on RAG integrated with prompt engineering as the strategy for adapting language models to specific science domains.}
\label{fig:spiderweb}
\end{figure}

In this context, we utilize an AI method known as Retrieval-Augmented Generation (RAG)~\cite{RAG_paper}. This approach allows us to retrieve factual information from an external knowledge base, providing language models with accurate and up-to-date data to enhance their understanding. A comparative analysis for LLM optimization strategies: RAG, RLHF and fine-tuning, is illustrated in Figure~\ref{fig:spiderweb}. The assessment is conducted considering key features for a workflow, as explained in Table \ref{tab:features}. These features encompass a range of issues, such as source-citing, hallucination, ease of changing and updating information, privacy and degree of personalization. The evolving nature of the field requires the consideration of features, such as ``Customization," which is particularly relevant due to the advent of manageable foundation models, such as LlaMA~\cite{touvron2023llama1}. These models pave the way for the potential development of intermediate-sized models in the future. 

Fine-tuning and RLHF strategies expose the model to several issues, such as hallucination and ``The Reversal Curse"~\cite{berglund2023reversal}. Additionally, they restrict the user to build on top of the initial model when fresh data is available and modifying the foundational model necessitates the arduous process of retraining. Hence, scoring for each strategy is derived based on extensive literature survey~\cite{ling2023domain, lialin2023scaling, rafailov2023direct, deci-2023, gudibande2023false} and careful consideration of such relevant factors, with the ideal scenario being perfect score in each category. From this analysis, prompt engineering provides a broader coverage of the key metrics for successful foundational model development. For this purpose, this strategy is further explored in this work.

\begin{table}[hbt!]
\centering
\begin{tabular}{lll}
\hline
\rowcolor[HTML]{EFEFEF} 
\multicolumn{1}{c}{\cellcolor[HTML]{EFEFEF}\textbf{No.}} &
  \multicolumn{1}{c}{\cellcolor[HTML]{EFEFEF}\textbf{Features}} &
  \multicolumn{1}{c}{\cellcolor[HTML]{EFEFEF}\textbf{Description}} \\ \hline
1 & Accuracy               & \footnotesize{Response accuracy for a given prompt}               \\
2 & Data Adaptability      & \footnotesize{Ease of inclusion of new datasets on regular basis} \\
3 & Hardware Adaptability  & \footnotesize{Heterogeneous hardware training and inference}        \\
4 & Energy Efficacy & \footnotesize{Total energy required to implement the method}      \\
5 & Data Privacy           & \footnotesize{Exposure to external entities, such as internet}        \\
6 & Robustness             & \footnotesize{Susceptibility to errors, such as hallucination}        \\
7 & Customization          & \footnotesize{Ease of updating the base LLM in the method}       \\ \hline
\end{tabular}
\caption{Features for comparing different LLM optimization strategies for information extraction task.}
\label{tab:features}
\end{table}

To motivate the current study, the state of foundational model representation in existing LLM frameworks is first presented. For this purpose, the topic of oblique detonation waves (ODWs) is used. Detonation-driven propulsion and energy conversion has become a recent topic of interest due to the higher efficiency of the underlying thermodynamic processes \cite{wolanski,ramanPrakash}. A variety of detonation approaches are being explored, including the use of oblique waves formed on a surface \cite{kareem_pnas,li-1995}. There exists a comprehensive literature on the structure of such waves spanning more than three decades, which forms the corpus of work used for the current study (more details provided in~\ref{odwList}). Using this topic, a short interaction with ChatGPT [September 25 2023 version] with different models and plugins is presented in Fig.~\ref{snippet-1chatGPT}. The response to first part of the question pertaining to definition of oblique detonation waves was satisfactory from all the models. On analyzing the response from GPT-3.5 model in box B, the paper with given title and \textbf{DOI: 10.1016/j.combustflame.2009.09.005} in the response is incorrect as it does not exist. This is a known problem of hallucination~\cite{hallucination} caused due to knowledge gap and model’s ability to generate seemingly credible text based on patterns it has acquired from its training data, regardless of whether the generated content conforms to actual facts. GPT-4.0's response in box C mentions the correct title of the article, however the DOI and publishing year are incorrect. Compared to GPT-3.5, the GPT-4.0 model is trained with more parameters and data. With additional plugin, as shown in box A, the model demonstrates the capability to generate a curated list of relevant research articles. These results consist of direct hyperlinks to the respective articles, reflecting a search engine-like functionality.

\begin{myboxVS}{\footnotesize{Question: What are oblique detonation waves? List relevant research articles}}\label{snippet-1chatGPT}
\begin{tcbraster}[raster equal height,raster valign=top, raster halign=center,raster columns=2, raster rows=1,
colframe=white,colback=white,colbacktitle=red!50!white,
]
\begin{tcolorbox}[title = A. GPT-4.0 + BrowserOp Plugin,colback=red!5!white,colframe=red!75!black,fonttitle=\bfseries, width = 7cm]
\footnotesize{Here are some relevant research articles on oblique detonation waves:

1.Title:[The formation and development of oblique detonation wave with different chemical reaction models]
(https://www.sciencedirect.com/science/
\\article/pii/S1270963821004740)

Content: This article explores the formation and development of oblique detonation waves with different chemical reaction models, focusing on flow field structure and radicals formation.

2.Title:[Stabilized detonation for hypersonic propulsion | PNAS]
\\
(https://www.pnas.org/doi/
\\10.1073/pnas.2102244118)
Content: The study discusses the creation of a standing oblique detonation wave...}
\end{tcolorbox}
\begin{tcolorbox}[boxsep=0pt,boxrule=0pt,colback=white,colframe=white,enhanced jigsaw,left=0mm,right=0mm,top=0mm,bottom=0mm]
\begin{tcolorbox}[title = B. GPT-3.5,colback=red!5!white,colframe=red!75!black,fonttitle=\bfseries, width = 7.85cm]
Title:\textcolor{red}{"Experimental and numerical study on the structure of oblique detonation waves in a stoichiometric hydrogen-air mixture"}
Authors: M. A. Ghotbi, M. Ashjaee, and K. J. Chung
Published in: Combustion and Flame, Volume 157, Issue 3, March 2010
\textcolor{red}{DOI: 10.1016/j.combustflame.2009.09.005}

\end{tcolorbox}
\begin{tcolorbox}[title = C. GPT-4.0,colback=red!5!white,colframe=red!75!black,fonttitle=\bfseries,  width = 7.85cm,before=\par\vspace{-0.2\baselineskip}]
Dolling, D. S. (\textcolor{red}{1984}). "Fifty years of shock-wave/boundary-layer interaction research: what next?." \textcolor{red}{AIAA journal, 22(9), 1228-1239}.
- This paper provides an overview of shock-wave interactions which are fundamental to understanding ODWs.

\end{tcolorbox}
\end{tcolorbox}
\end{tcbraster}
\end{myboxVS}

These responses provide insight into current limitations. First, LLMs are phrase-completion models that may not directly yield accurate responses. Second, the corpus of data used in the training is not transparently available to the user. Third, the responses have to be verified, for instance, through checks for existence of DOI links. Given these limitations, identifying pathways to increasing the response accuracy of the model is an important step. Short of training foundational models from scratch, the best strategy (see Fig.~\ref{fig:spiderweb}) is to develop optimal strategies for querying the model, and by providing relevant data for informed outputs. In this work, this prompt optimization strategy is formulated and demonstrated using the topic of ODWs.

\section{Method}
\begin{figure}[hbt!]
\centering
\includegraphics[width=1.\textwidth]{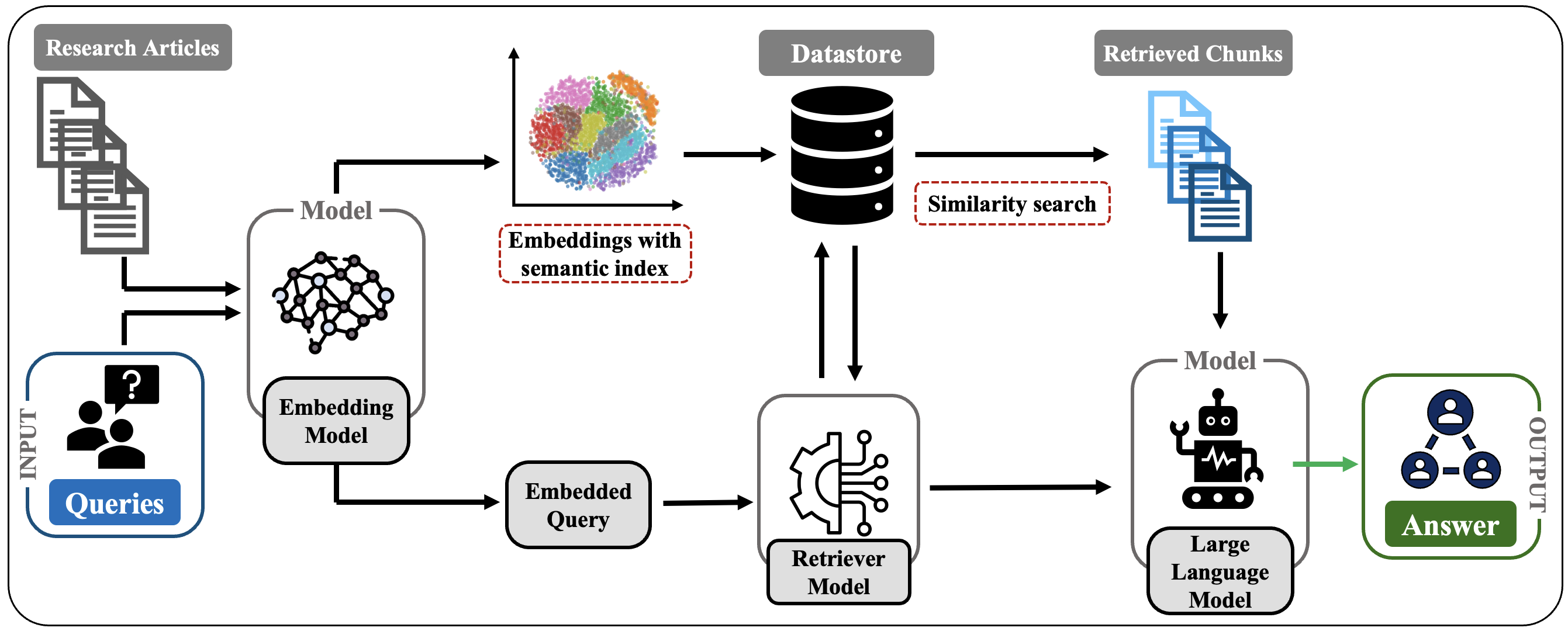}
\caption{Process workflow for information retrieval and querying.}
\label{fig:wrkflow}
\end{figure}

The workflow used in this work is illustrated in Fig.~\ref{fig:wrkflow}. Initially, a collection of PDFs is loaded into memory as binary data, subsequently transformed into textual content, which is then subjected to an embedding model to generate embeddings. These embeddings are stored locally in a vectorized database. Once this phase is complete, the LLM can be initialized, enabling users to query the database through a predefined prompt template. Selecting the appropriate open-source LLM, such as LlaMA-1 family~\cite{touvron2023llama1}, LlaMA-2 family~\cite{touvron2023llama2} or various related variants, requires taking into account several key factors. This discussion is beyond the scope of this work, and a model with sufficient large token limit was selected from Open LLM Leaderboard~\cite{open-llm-leaderboard, eval-harness, clark2018think, zellers2019hellaswag, hendrycks2021measuring, lin2022truthfulqa}. A large token limit enables the model to efficiently process substantial number of embeddings within a single interaction. This concept draws a parallel with the fidelity level of a computational fluid dynamics (CFD) solver, wherein a higher fidelity configuration can accurately capture finer details of the physics involved. The selected model is quantized~\cite{jacob2017quantization} to run on CPUs, thereby reducing memory footprint and improving inference speed and efficiency. Nevertheless, a trade-off exists between output precision and computational capacity. Users are provided with the flexibility to choose the quantization level that best suits their requirements in order to achieve optimal performance. The subsequent sections provide a detailed description of each sub-task within the workflow.

\subsection{Document Ingestion and Database Creation}

\begin{figure}[hbt!]
\centering
\includegraphics[width=1.\textwidth]{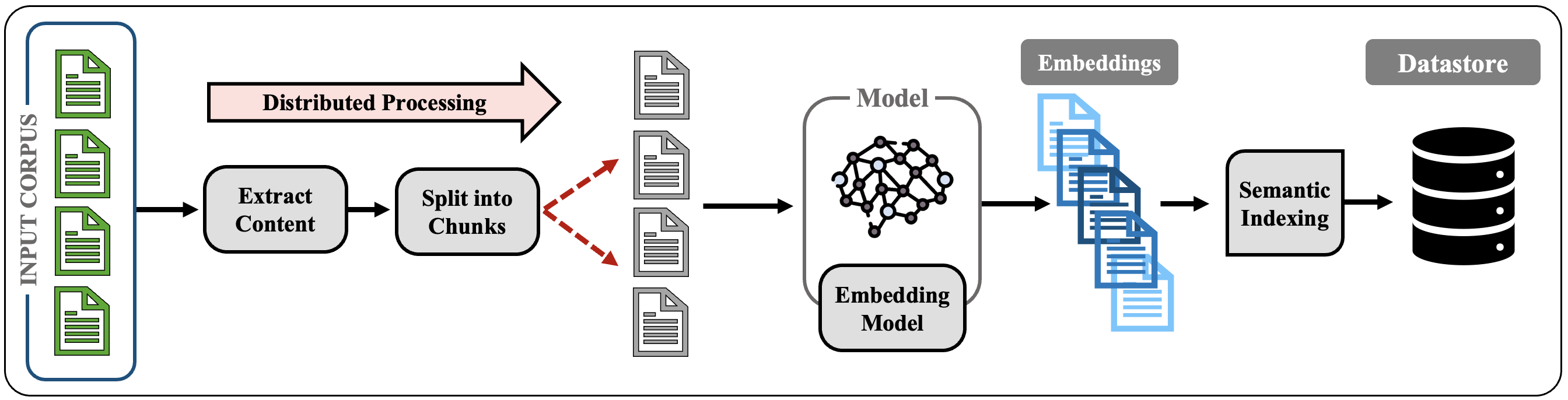}
\caption{Embedding documents using multi-processing framework to persist in a database. }
\label{fig:pdf_load}
\end{figure}

The database creation pipline is shown in Fig.~\ref{fig:pdf_load}. The workflow begins with storing PDFs locally on the device. To load the documents for database creation the open-source library LangChain~\cite{langChain} is used. This library provides specialized tools or chains tailored to create applications powered by LLMs. The \codebox{UnstructuredPDFLoader} function coupled with \codebox{DirectoryLoader} is used to load the document data in binary format. The loading procedure can be expedited by leveraging the multi-threading feature. Subsequently, the imported unstructured document data is partitioned into  specific chunks of user-defined size. These chunks are further converted into embeddings using a transformer-based model, with the option to choose from a range of models accessible on the Massive Text Embedding Benchmark (MTEB) Leaderboard~\cite{muennighoff2022mteb}. In pursuit of optimizing the embedding procedure, we have integrated the Ray framework~\cite{ray} into our workflow. Ray provides valuable functionalities (for example \codebox{ActorPoolStrategy}) aimed towards facilitating parallelization of the embedding process through GPU utilization. The output of the embedding process are encoded vectors or arrays of numbers that represent the meaning and the context of the input chunks. These arrays help the LLM to understand the semantic and syntactic relationships between the input text, and to generate more relevant and coherent responses. Therefore, understanding text splitting and chunking strategy is critical for knowledge extraction and querying process (described further in section~\ref{cSize}). 

The embeddings are retained within an indexed vector database, which provides a knowledge extension of the LLM and querying capabilities tailored to the unique structure of vector embeddings. Some of the frequently used open source databases are Chroma~\cite{chroma}, Faiss~\cite{johnson2019billion_faiss}, Qdrant~\cite{qdrant-no-date} and Vespa~\cite{vespa}. With multiple user queries to the same data, databases act as a long-term LLM memory and can cache previous interactions between the model and the user. This capacity serves to mitigate LLM response latency and effectively overcomes the limitations related to context length or token constraints inherent to the model. It is noteworthy that the present study exclusively addresses the retrieval of contextual information from the knowledge database, which is based on Chroma \cite{chroma}.

\subsection{Querying and generative agent} \label{llm_query}
\begin{figure}[hbt!]
\centering
\includegraphics[width=1.\textwidth]{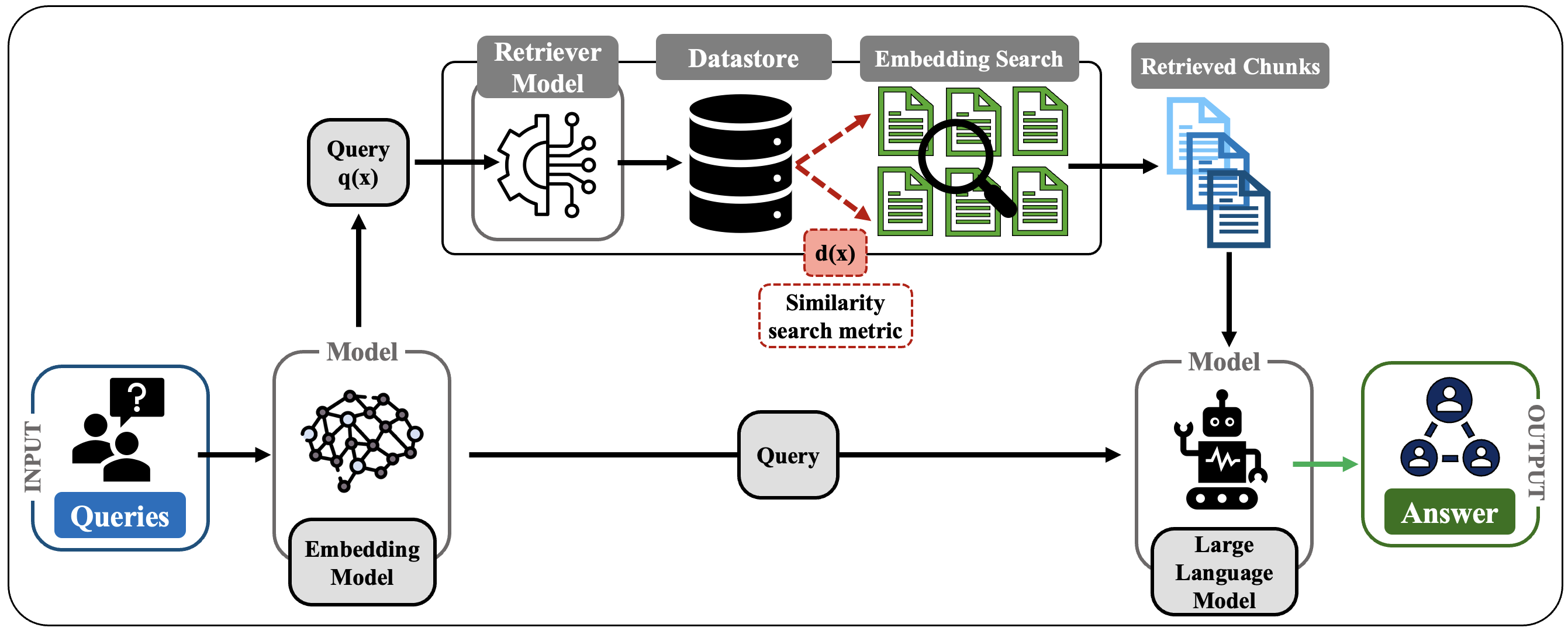}
\caption{Workflow for generative answering process.}
\label{fig:wrkflow_LLM}
\end{figure}

Interaction between different components such as user queries, LLM and knowledge database occurs in a single pipeline known as ``chain". A detailed representation of this pipeline is shown in Fig.~\ref{fig:wrkflow_LLM}.  
The user interacts with the pipeline by submitting a query $x$, which is converted into an embedding, denoted as $q(x)$, using the same embedding model ``$q$" that was used during the database creation process. The context retriever constitutes an algorithm designed to streamline the extraction of embeddings closely associated with the query embedding $q(x)$. This can be achieved through two primary approaches: a) Semantic Similarity, and b) Maximum Marginal Relevance (MMR) as detailed in~\cite{langchain_MMR} and mathematically expressed as shown below.
\begin{equation}
M M R \stackrel{\text { def }}{=} \operatorname{Arg} \max _{D_i \in R \backslash S}\left[\lambda\left(\operatorname{Sim}_1\left(D_i, Q\right)-(1-\lambda) \max _{D_j \in S} \operatorname{Sim}_2\left(D_i, D_j\right)\right)\right]
\end{equation}
where, $Sim_1$ and $Sim_2$ are similarity metrics (Table~\ref{tab:distanceMetric}), 
$Q$ = Query (description of document category), 
$D$ = Set of documents related to Query $Q$, 
$S$ = Subset of documents in $R$ already selected,  
$R \backslash S$ = set of un-selected documents in $R$, and  
$\lambda$ = Constant in range [0–1], for diversification of results.
 
In a basic semantic similarity search, the method identifies and selects embeddings that exhibit the highest degree of similarity to the provided inputs. On the other hand, the MMR approach takes a different route by initially filtering embeddings with the greatest similarity to the inputs. Subsequently, MMR iteratively incorporates these filtered embeddings, while simultaneously penalizing them for their proximity to previously selected examples. MMR, as a crucial technique, plays a pivotal role in promoting diversity within search results. On the contrary, the semantic similarity approach inadvertently excludes a broader range of diverse information.

Measuring similarity plays a crucial role in information retrieval process. In order to measure similarity between query ($q(x)$) and embedded data, the database applies algorithms to draw commonalities between the two embedding patterns. At the most basic level, for given two embedding vectors $X = [x_1, x_2, ..., x_n]$ and $Y = [y_1, y_2, ..., y_n] \in \mathbb{R}^{n}$, similarity measures can be expressed as shown in Table~\ref{tab:distanceMetric}. These distance-measure formulations cannot be directly applied to large datasets due to computation complexity added by an exhaustive search, where distance between each vector is computed. To efficiently search similar objects at a large-scale, supervised learning algorithms, such as $k$-NN or clustering algorithms, such as K-means are implemented that use the measures in Table~\ref{tab:distanceMetric}. We use cosine similarity measure as it can accommodate variable-length data. In scenarios with two similar data vectors exhibiting significant separation in Euclidean distance owing to their size disparities, this metric can still reveal their relatively small angular discrepancy. It is crucial to note that a smaller angular disparity signifies a stronger similarity between these data vectors.

\begin{table}[hbt!]\setlength{\tabcolsep}{2.5pt}
\centering
\begin{tabular}{llr} 
\hline
\rowcolor[HTML]{EFEFEF} 
\multicolumn{1}{c}{\cellcolor[HTML]{EFEFEF}\textbf{Metric}} &
\multicolumn{1}{c}{\cellcolor[HTML]{EFEFEF}\textbf{Description}} &
\multicolumn{1}{c}{\cellcolor[HTML]{EFEFEF}\textbf{Formula}} \\ \hline
\\[-1em]
\footnotesize{Minkowski Distance} & \footnotesize{Distance between vector in Normed vector space}       & $\left(\sum_{i=1}^n\left|x_i-y_i\right|^p\right)^{\frac{1}{p}}$ \\
\footnotesize{Euclidean Distance} & \footnotesize{Distance between ends of vectors}                     & $\left(\sum_{i=1}^n\left|x_i-y_i\right|^{p=2}\right)^{\frac{1}{p=2}}$ \\
\footnotesize{Manhattan Distance} & \footnotesize{Distance between two data points in a grid-like path} & $\left(\sum_{i=1}^n\left|x_i-y_i\right|^{p=1}\right)^{\frac{1}{p=1}}$ \\
\footnotesize{Chebychev Distance} & \begin{tabular}[c]{@{}l@{}}\footnotesize{Distance between two vectors is determined by} \\ \footnotesize{the max difference along any coordinate dimension}\end{tabular} &
  $\left(\sum_{i=1}^n\left|x_i-y_i\right|^{p\to \infty}\right)^{\frac{1}{p\to \infty}}$ \\
\footnotesize{Cosine}             & \footnotesize{Cosine of angle $\theta$ between vectors}             & $\frac{\sum_{i=1}^n x_i y_i}{\sqrt{\sum_{i=1}^n x_i^2} \cdot \sqrt{\sum_{i=1}^n y_i^2}}$ \\
\footnotesize{Inner Product}      & \footnotesize{Cosine multiplied by lengths of both vectors}         & $\sum_{i=1}^n x_i y_i = |x||y|\cos(\theta)$ \\ \hline
\end{tabular}
\caption{Vector search similarity metrics used in databases for information retrieval.}
\label{tab:distanceMetric}
\end{table}

\subsection{Optimal User Prompts}
\begin{figure}[hbt!]
\centering
\includegraphics[width=0.55\textwidth]{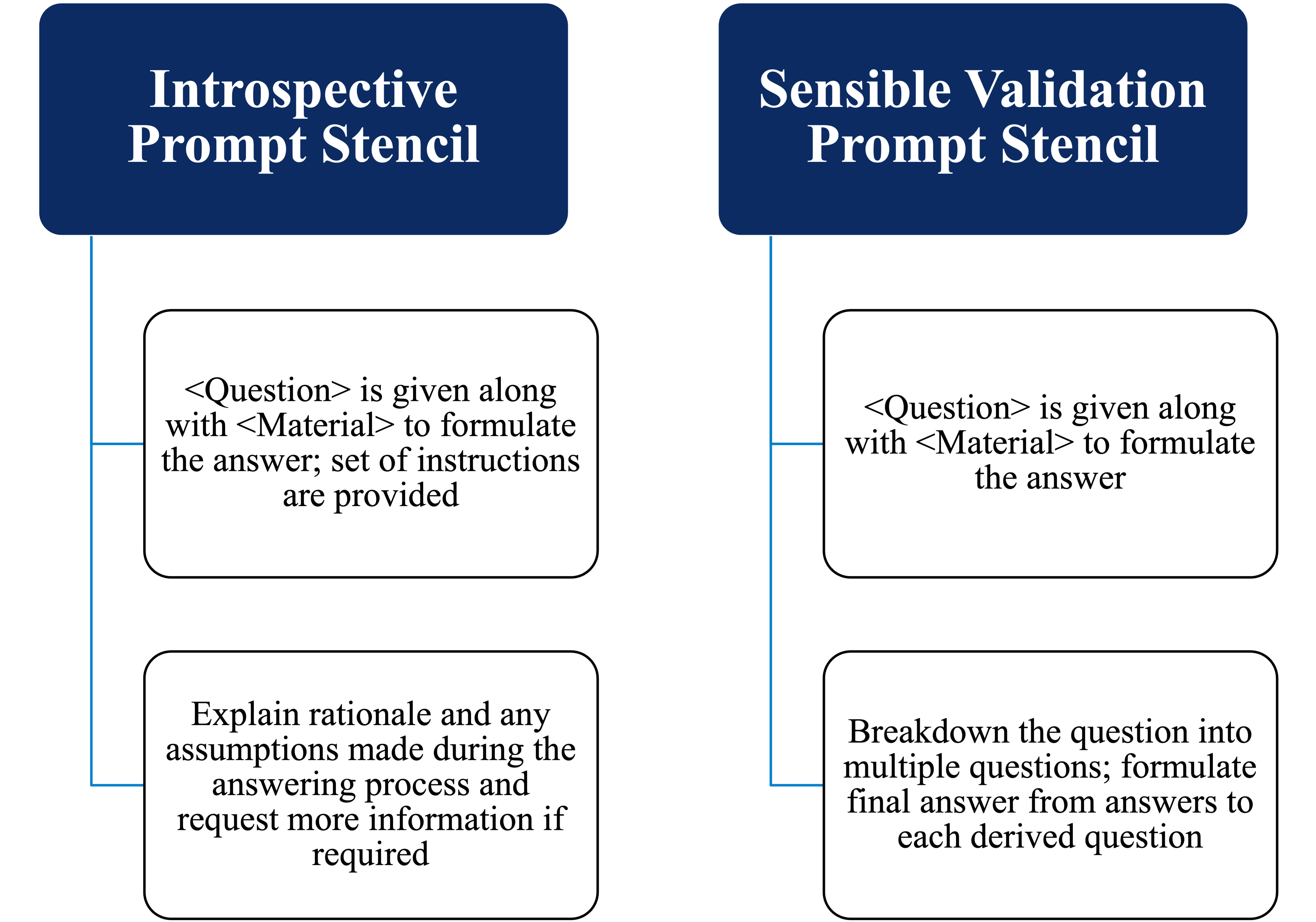}
\caption{Optimal prompt stencils structures for knowledge extraction. }
\label{fig:prompt_LLM}
\end{figure}

Interaction with LLMs is strongly influenced by user prompts as they provided structured set of instructions or contextual information pattern to the model. In the context of our current study, it is important to establish an optimal prompt pattern that empowers the LLM to intricately grasp the underlying intent and context when generating responses. To achieve this, two stencil strategies are proposed here (Fig.~\ref{fig:prompt_LLM}): the Introspective Prompt Stencil and the Sensible Validation Prompt Stencil. These stencils incorporate questions (\textit{Question}) and text snippets derived from research papers within our database (\textit{Material}), serving as the foundation for our structured interactions with the LLM.  

In introspective stencil, directive to the LLM involves the dual tasks of presenting a clear rationale for the generated response, as well as uncovering any information gaps in the response. This enables the model to reveal any implicit assumptions embedded within the answers and thereby aids users in identifying potential ambiguities, especially when dealing with subjects that exhibit significant knowledge overlap. 

The sensible validation stencil instructs the LLM to divide the question into discrete sub-queries, and formulate the final response by combining answers to these sub-queries. This strategy helps LLM to grasp the underlying context and facilitates the identification of any absent details. By combining the answers to sub-queries, the final answer becomes comprehensive and, in a sense, guarantees the inclusion of all pertinent information in the response. However, it is important for the user to judiciously manage the generation of invaluable sub-questions by imposing a limit on the number of questions that can be generated. In this study, the prompt stencil is limited to a combination of these two strategies, and for additional information the readers are directed to~\cite{white-2023, gao-2023}. 

\section{Results}
In this section, the framework as applied to the ODW dataset is investigated in detail. All results are produced by a single language model with variations in database characteristics and prompts. The \textit{bge-base-en-v1.5} model from BAAI~\cite{bge_embedding} is used for embedding the data segments or chunks. The choice of this model is predicated on its embedding vector dimensionality of 768, aligning with the specifications of the LlaMa-2 7B model. 

\subsection{Effect of Chunk Size}\label{cSize}
In the text segmentation procedure, a recursive approach~\cite{recursiveTextSplit, casey-1996, chrupala2013text} is used to divide the text into smaller, hierarchically organized chunks utilizing a designated set of separators. If the initial attempt to split the text does not yield segments of the desired size or structure, the algorithm invokes itself recursively on the resultant segments, employing alternative separators or criteria, until the desired chunk size or structure is attained. This recursive text splitting strategy enables the effective segmentation of text into meaningful units.
The two critical parameters in the procedure are \textbf{chunk-size} and \textbf{chunk-overlap} that control the segment size and the degree of correlation maintained between successive segments, respectively.

To study the effect of these parameters, five research articles are selected from the corpus (~\ref{odwList}), and the embeddings analyzed using T-distributed stochastic neighbor embedding (t-SNE) tool~\cite{tSNE}. t-SNE is particularly useful for reducing the dimensionality of the embedding space, facilitating visualization in the form of clusters. In t-SNE plots, each point represents an embedding vector, and the clusters collectively represent all vectors associated with a specific entity—in this context, a research article.

\begin{figure}[hbt!]
\centering
\includegraphics[width=1.\textwidth]{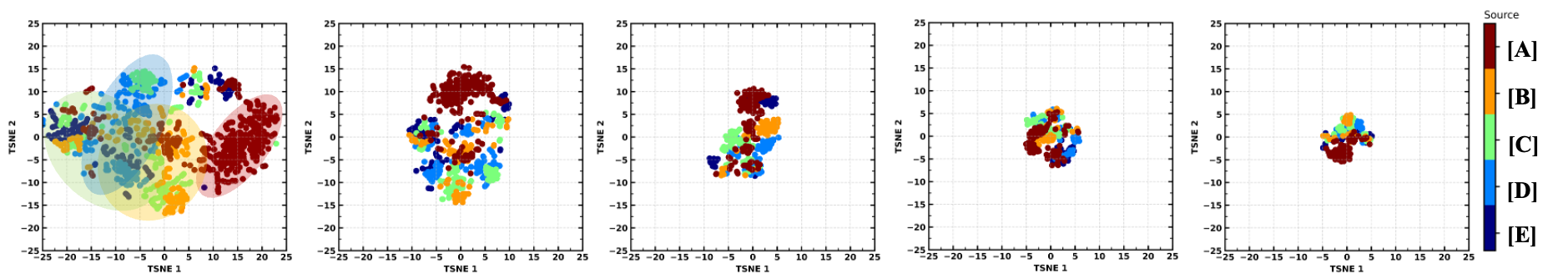}
\caption{Effect of increasing chunk size (left to right) during text embedding process.}
\label{fig:chunkSize}
\end{figure}

Figure~\ref{fig:chunkSize} depicts the effect of increase in chunk size on embedding clusters while holding the chunk overlap constant. The chunk size and, consequently, the array size based on character length are increased from 800 to 2000 in increments of 300, maintaining a constant overlap of 500. This progression is illustrated in plots arranged from left to right. An approximate location and spread of the clusters of the different sources are marked in the left image. Intuitively, it can be seen that as the chunk size is increased, with constant overlap, the clusters shrink. The phenomenon is ascribed to the impact of increasing text within each segment. Since the quantity of data remains constant, the elongation of data length in each vector occurs, consequently diminishing the count of cluster points and augmenting the information encapsulated in each point.
To further explore this aspect, three databases using the chunk size of 800 (left), 1200 (middle), and 2000 (right) are considered. The following three questions were then posed and the results made available to four domain experts, who rated the answers on a scale of 1-5: Q-1) \textit{what is an oblique detonation wave?} Q-2) \textit{How does shock boundary layer interaction affect oblique detonation wave structure?} Q-3) \textit{Do shear layers affect oblique detonation wave initiation structure?}

\begin{figure}[hbt!]
\centering
\includegraphics[width=0.45\textwidth]{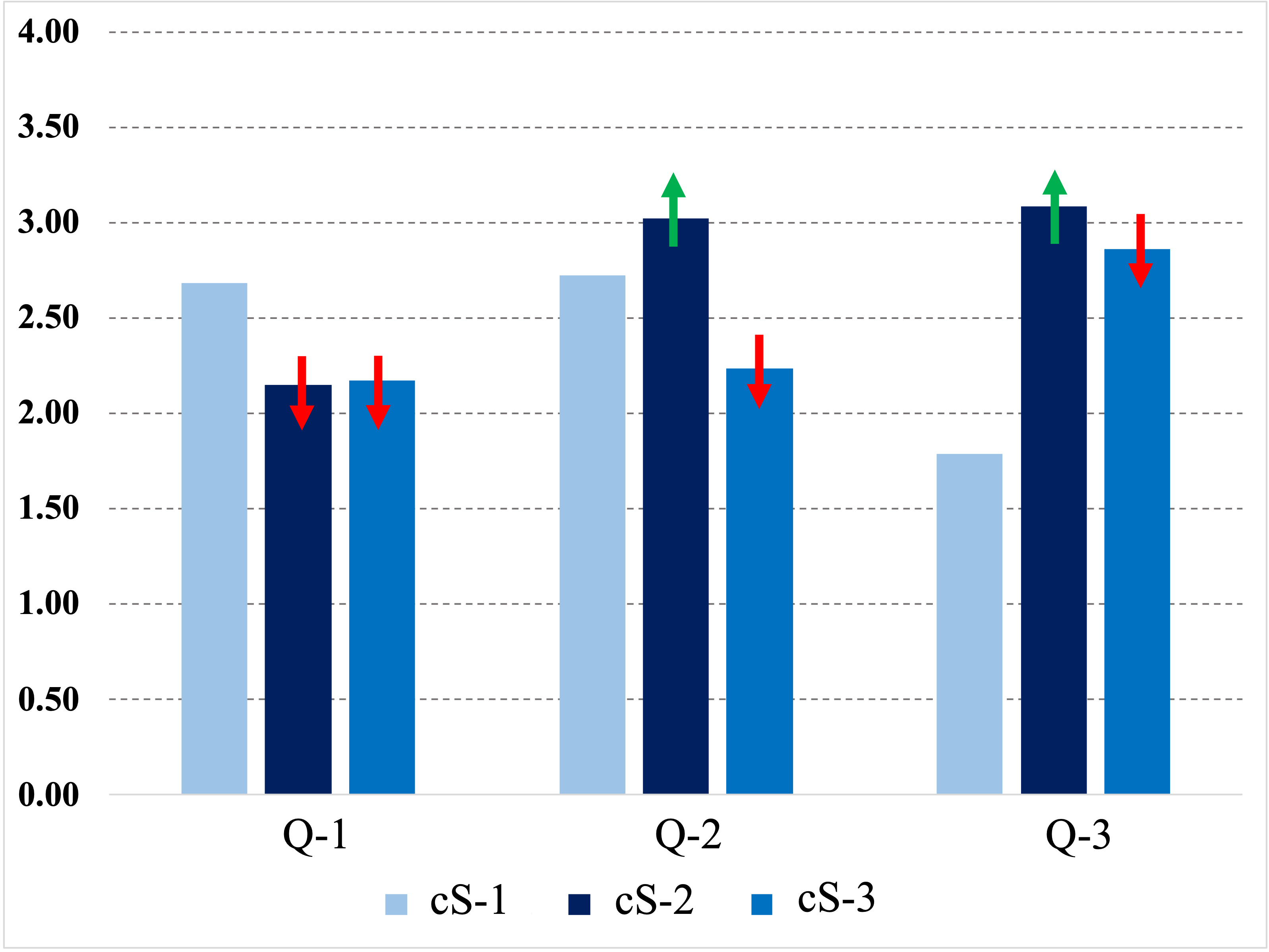}
\caption{Evaluation scores for database tests using subject-specific questions to determine optimal chunk size.}
\label{fig:chunkSizeRes}
\end{figure}

The evaluated mean scores are shown in Fig.~\ref{fig:chunkSizeRes}, where ``cS" refers to the database in order 800, 1200 and 2000 character length embeddings. The arrows mark relative change of scores in each question's bar plot only. As seen for question Q-1, cS-1 performs marginally better compared to other databases. However, in the remaining two questions, the cS-2 dataset emerges as the highest scorer with a good margin from both the models at different occasions. This can be related to decrement of inter-clusters distance in Fig.~\ref{fig:chunkSize}, where the right-most image for chunk size of 2000 concentrates the clusters towards the center. As the size of the data segments expanded, the overlap between the segments was held constant that led to a diminished contextual coherence between successive segments. In this context, contextual coherence refers to the degree of similarity between successive data units. For instance, two successive paragraphs in a research article may or may not exhibit coherence; however, two sentences within a paragraph are more likely to be coherent. Conversely, this phenomenon was not observed in the case of smaller data segments, where an equilibrium appeared to exist between the informational content encapsulated within each segment and its contextual overlap with adjacent segments. This balance provided the language model a more comprehensive dataset, thereby enhancing its capacity to generate responses of greater breadth and contextual relevance.

\subsection{Effect of Chunk Overlap}
\begin{figure}[hbt!]
\centering
\includegraphics[width=1.\textwidth]{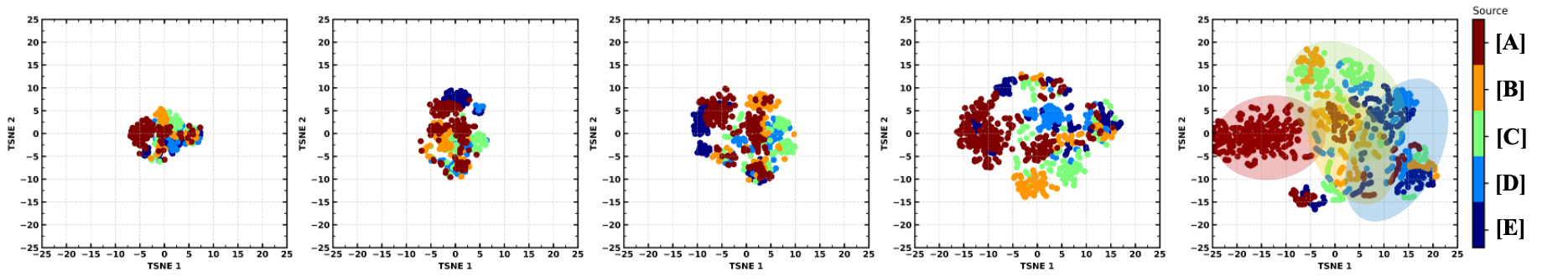}
\caption{Effect of increasing chunk overlap (left to right) during text embedding process.}
\label{fig:chunkOverlap}
\end{figure}

A similar approach is applied to study effects of chunk overlap on contextual data in each segment during the embedding process. The Fig.~\ref{fig:chunkOverlap} shows t-SNE output for increasing chunk overlap (left to right) for a fixed chunk size. The overlap is varied from 0 to 700 in steps of 175, with a fixed chunk size of 1000. 
The plots reveals a clear tendency for clusters to segregate, accompanied with the emergence of new data points within each cluster. Notably, the inter-cluster distance demonstrates an increasing trend, accompanied by a marginal increment in intra-cluster scatter. To conduct a more comprehensive analysis, a similar strategy to section~\ref{cSize} is employed herein. Three distinct databases are constructed, characterized by chunk overlaps of 0 (left), 350 (middle), and 700 (right), respectively. Subsequently, these databases are then subjected to evaluation based on the three aforementioned questions by subject experts.
\begin{figure}[hbt!]
\centering
\includegraphics[width=0.45\textwidth]{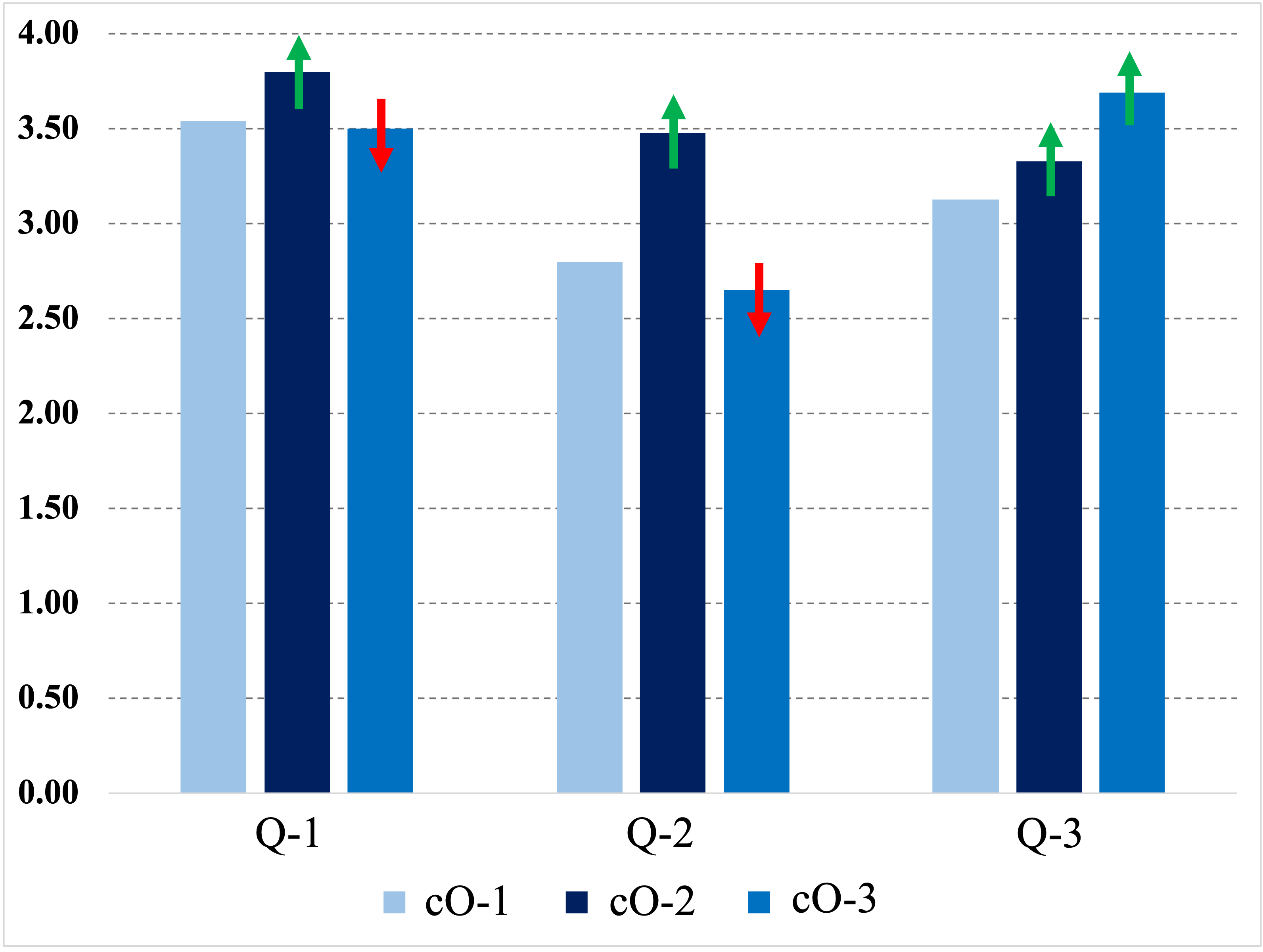}
\caption{Evaluation scores for database tests using subject-specific questions to determine optimal chunk overlap.}
\label{fig:chunkOverlapRes}
\end{figure}

The findings presented in Figure~\ref{fig:chunkOverlapRes} depict the mean scores corresponding to each database denoted by ``cO." Notably, for questions Q-1 and Q-2, the database characterized by a chunk overlap of 350 demonstrated the highest scores, displaying substantial margin on multiple occasions. However, in the case of Q-3, the database with an overlap of 700 exhibited the highest score, albeit with a margin less pronounced than observed in the aforementioned questions. It was obvious that cO-1 database with no overlap will not perform exceptionally well compared to the other databases. It is noteworthy that the assessment scores presented in Fig.~\ref{fig:chunkSizeRes} exhibit, on average, lower values in comparison to the scores depicted in Fig.~\ref{fig:chunkOverlapRes}. This observation suggests that, within the context of scientific knowledge extraction, the impact of chunk overlap may surpass that of chunk size. This evaluation will be further explored in the next paragraph.
Furthermore, these scores can be linked to plots seen in Fig.~\ref{fig:chunkOverlap}, where the right plot for 700 overlap has clear clusters segmented and scattered apart showing the data segments have increased contextual coherence between successive segments. However, with a fixed chunk size of 1000 and overlap of 700, results in 70\% of the data segment potentially sharing context with other segments leading to reduction in actual informational content within the segment itself. The database cO-2 presents a balance between contextual overlap and information stored within the segment.

At this point, the analysis for chunk size and overlap clearly highlights the pivotal role played by the length of the input character array in shaping the model's ability to assimilate contextual information and generate a final response.
In this context, the term "tokens" can be introduced without causing confusion with chunks, which are character arrays. Tokens represent segmented individual words or other meaningful units, such as phrases or named entities. Consequently, a token can encompass multiple chunks and can be regarded as a fundamental unit in NLP. Each LLM is constrained by a token limit, dictating the volume of information the model can process at once based on its architecture. Specifically, the LlaMa-2 7B model has a token limit of 4096, and the embedding model (EM) is restricted to 768 tokens. 

\begin{figure}[hbt!]
\centering
\includegraphics[width=1.0\textwidth]{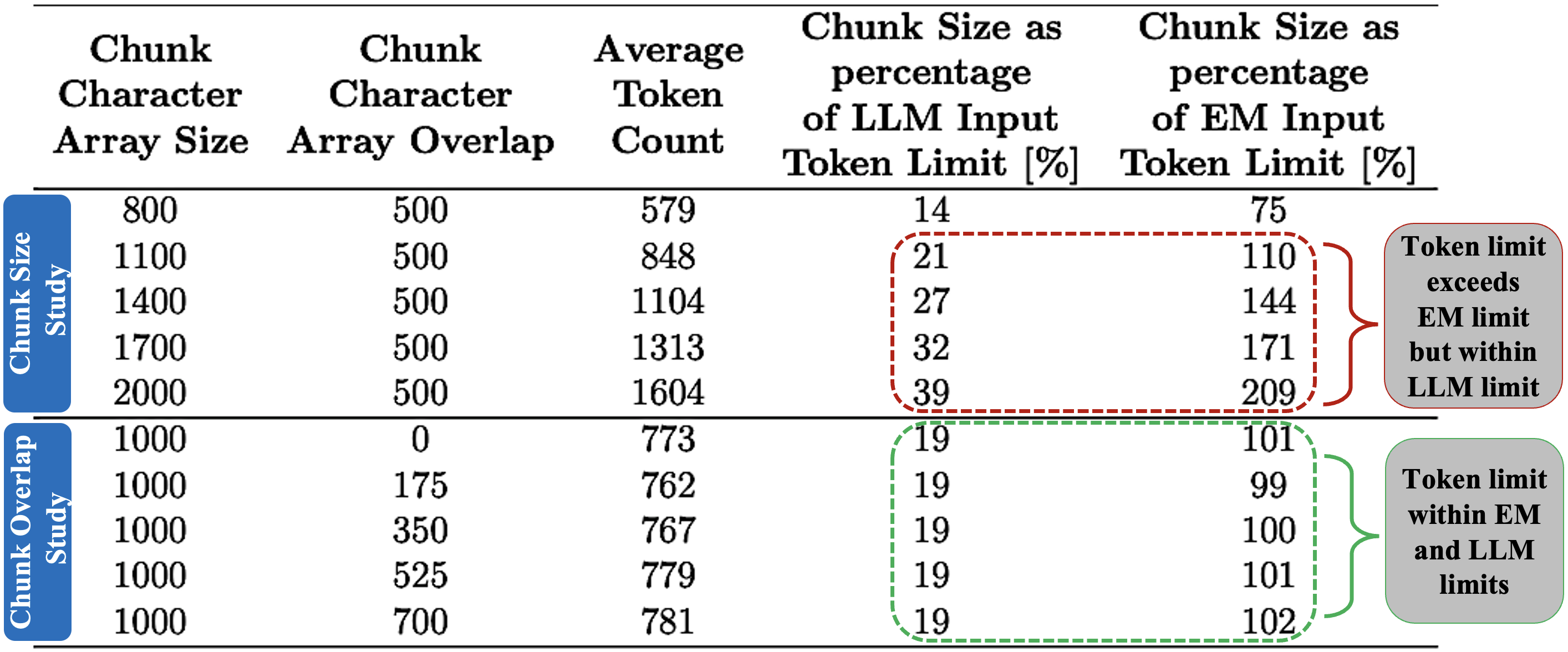}
\caption{Table for token limit evaluation based on LLM and embedding model input token limits.}
\label{fig:tableTokens}
\end{figure}
The table shown in Fig.~\ref{fig:tableTokens} displays the token counts for the chunk size and overlap studies described earlier. Subsequently, the token count for each embedding vector is represented as a percentage of both the LLM's token limit and the EM's token limit. 
For the chunk size study, the token size increased as the array size was incremented. Based on the percentages of the maximum limits, the 1600 array size was merely 40\% of the LLM's limit, but for the EM, it was 200\% bigger than what the model could process. This may result in a loss of precision in information and introduce noise in the data chunks. The better approach will be to match the EM's token limit, as seen in the second row for the chunk overlap study, where the chunk size percentage is around 100\%. As mentioned before and evident in the evaluation scores (Fig.~\ref{fig:chunkOverlapRes}), the ``cO"-databases used for Q-A scored higher than the ``cS"-databases. Clearly, the reason was data chunks in the cO-database were less noisy and captured information better. In the overlap study, the chunk size constitutes only about 20\% of the LLM limit. This allows for the simultaneous submission of multiple chunks to the LLM, facilitating the generation of responses with rich and diverse arguments.

The insights presented in these sections are crucial for developing a strategic approach to determine optimal chunk size and overlap parameters. It is advisable to set the overlap ratio within the range of 20\% to 40\% relative to the chunk size. The selection of the chunk size is dependent on the embedding model and should fall within the range of 85\% to 95\% of its dimension.

\subsection{Demonstration of Knowledge Extraction}

In this section, the language model framework (Fig.~\ref{fig:wrkflow}) is used to compile a set of responses for questions relevant to ODW research. The analysis encompasses an evaluation of the key information and comprehensiveness of the responses generated by the model.

\subsubsection{Vanilla Model vs Framework based Model}
First, the responses from a standard/vanilla LLM (vLLM) are compared with those from a Framework based LLM (fLLM). In this context, ``vanilla LLM" denotes a basic language model that operates without any supplementary database access, relying solely on its pre-trained knowledge base. On the other hand, ``Framework LLM" refers to the vanilla LLM integrated with the framework outlined in Figure~\ref{fig:wrkflow}. This analysis helps us comprehend improvements solely due to the approach taken in this study. The prompt used for this analysis is \textit{``Reply with minimum 100 words and provide give a detailed list of research papers for this topic. If you don't know the answer, just say that you don't know, don't try to make up an answer"}. The context was provided using text snippets retrieved from the database in the case of framework only. 

The models were subjected to three queries, as delineated in the ensuing snippets along with an additional prompt. We will elaborate on one query here, and for the remaining queries, please refer to~\ref{vanillaLLMQA}. Evidently, the responses generated by the fLLM exhibit a greater level of verbosity upon initial inspection when contrasted with the responses from the vLLM. The explanations provided by the vLLM to the fundamental questions seem more comprehensive initially, as exemplified in the Q2 snippet where the vLLM expounds on the impact of shear layers for a span of 3-4 lines before presenting a list of pertinent research articles to the user. In contrast, the fLLM does not consistently furnish a complete description within the confines of 3-4 lines; instead, it explains key points to formulate responses while concurrently presenting an extensive list of articles explaining the work conducted in those specific publications. The citation approach employed by the fLLM possesses a distinctive pattern: it commences by highlighting the study's objectives, followed by the articulation of key findings pertinent to the question's core content. Consequently, it furnishes a comprehensive response to the queries, occasionally resembling the style of a researcher well-versed in the domain of the question's subject matter.

An important aspect of the question was to provide a list of coherent research articles to the user. The vLLM provided a catalog of article that do not exist. The authors are fictitious and search through diverse databases failed to reveal any actual author associated with an article featuring even partially congruent keywords from the title of the presented paper in the response. This behaviour is consistent with different variations of chatGPT illustrated in snippet~\ref{snippet-1chatGPT}. 
In contrast, the fLLM system demonstrates a more comprehensive listing accompanied by citations, as exemplified by \textcolor{red!65}{\textit{Li, Kailasanath \& Oran (1994): "Oblique Detonation Waves in Wedge Flows." Combustion Science and Technology, 96(1), 57-73}}, within the context of Q1. Although an article sharing an identical title does not exist, it is noteworthy that the authors collaborated on a paper titled \textcolor{blue!65}{\textit{``Detonation Structures Behind Oblique Shocks"}[https://doi.org/10.1063/1.868273]} in the same year, 1994. To address this discrepancy, a thorough analysis of the data fragments retrieved from the database and supplied to the model was conducted. It was observed that one of the data chunks contained the reference: \textit{``Li, Kailasanath \& Oran (1994) later simulated the oblique detonation and described"}, which was utilized by the fLLM during the formulation of its response. This discovery holds significant importance, shedding light on the manner in which the LLM attempted to address the query with limited context. When the LLM attempted to reconstruct the missing details of the referenced article based on the adjacent text, it generated a fictional title. Notably, the model anticipated the content of the article, correctly identifying its potential focus on oblique detonation waves in wedges. Furthermore, the model incorrectly projected the journal's issue year as 1996, despite being aware that the article was published in 1994. It should be noted that this assessment may vary if the data fragments lack author names, presenting only citation numbers from the bibliography.

This addresses a pivotal question:~\textbf{\textit{Does sentence completion serve as the operational mechanism for the scientific knowledge extraction task at hand?}}. Examining the received responses, there is a plausible inference that the model engages in the systematic completion of sentences, effectively bridging knowledge gaps and rephrasing sentences predicated solely on the input data chunks, with minimal dependence on its internalized pre-existing knowledge. The next section presents strategies to mitigate this challenge. 

\begin{myboxVS}{Q1: what is an oblique detonation wave? + Prompt}
\begin{tcbraster}[raster equal height,raster valign=top, raster halign=center,raster columns=1, raster rows=2, colframe=white, colback=white,  ]

\begin{tcolorbox}[title = Vanilla LLM, colbacktitle=black!40!white,
colback=black!5!white,colframe=red!0!black, fonttitle=\bfseries,height=6.5cm]
\footnotesize{An oblique detonation wave is a type of detonation wave that occurs when a shock wave impinges on a flat or curved surface at an angle. This can occur in various applications such as explosions, rocket launches, and gas flows. The oblique detonation wave can produce complex and diverse wave patterns, including shock waves, pressure waves, and velocity waves, which can have significant effects on the surrounding material.
Research Papers:
1. "Oblique Detonation Waves" by J. S. Shen, Journal of Fluid Mechanics, 1975.
2. "Detonation Waves and Their Applications" by A. H. Stapleton, Annual Review of Fluid Mechanics, 1983.
3. "Oblique Detonation in a Channel with a Curved Wall" by M. Taheri, Journal of Fluid Mechanics, 2006.
4. "Numerical Simulation of Oblique Detonation Waves in a Two-Dimensional Rectangular Channel" by Y. Zhang, Journal of Computational Physics, 2013.
5. "Experimental Study on the Formation and Propagation of Oblique Detonation Waves in a Confined Channel" by X. Liu, Experimental Thermal and Fluid Science, 2019.}
\end{tcolorbox}

\begin{tcolorbox}[title = Framework+LLM, colbacktitle=blue!50!white, colback=black!5!white, colframe=red!0!black, fonttitle=\bfseries,height=12cm]
\footnotesize{Oblique detonation wave is a type of combustion wave that occurs when a supersonic ow interacts with a wedge or a cone. The oblique shock wave formed at the leading edge of the wedge or cone then undergoes a series of complex interactions, including reaction, heat transfer, and mass transfer, which ultimately lead to the formation of an oblique detonation wave. This type of combustion wave has been widely studied in recent years due to its potential applications in propulsion systems, such as oblique detonation wave engines (ODWEs) and Ram Accelerators (Nettleton, 2000).
Research Papers:
1. Li, Kailasanath \& Oran (1994): "Oblique Detonation Waves in Wedge Flows". Combustion Science and Technology, 96(1), 57-73.
2. Teng, H., Jiang, Z., \& Ng, H. D. (2016): "Supersonic Combustion Based on Oblique Detonation Waves". Progress in Aerospace Sciences, 98, 62-85.
3. Nettleton, P. (2000): "Oblique Detonation Wave Engines: Theory and Applications". Journal of Propulsion and Power, 16(4), 576-585.
4. Roy, R., Chaudhuri, S., \& Basu, S. (2004): "Oblique Detonation Waves in a Wedge: Experimental Studies". Combustion Science and Technology, 176(9), 1327-1346.
5. Kailasanath, B. (2003): "Oblique Detonation Waves in Supersonic Flows: A Review". International Journal of Advanced Manufacturing Technology, 21(7-8), 597-614.
6. Choi, J., Kim, D., \& Lee, S. (2013): "Simulation of Oblique Detonation Waves in a Wedge Using Large Eddy Simulation". Journal of Aerospace Engineering, 26(2), 209-224.
7. Zhang, Y., \& Tian, F. (2013): "Oblique Detonation Wave Formation in a Wedge: Numerical Study". Journal of Computational Physics, 258, 173-191.
8. Teng, H., Jiang, Z., \& Ng, H. D. (2017): "Oblique Detonation Wave Dynamics in a Wedge: An Experimental Study". Combustion Science and Technology, 189(6), 953-967.
9. Liu, Y., Wang, J., \& Zhang, J. (2018): "Oblique Detonation Wave Characteristics in a Wedge with Different Initiation Angles". Journal of Thermophysics and Heat Transfer, 32(4), 756-766.
10. Zhang, J., \& Tian, F. (2018): "Oblique Detonation Wave Dynamics in a Conical Wedge: Numerical Study". Journal of Computational Physics, 369, 144-157}
\end{tcolorbox}

\end{tcbraster}
\end{myboxVS}

\subsubsection{Optimized Prompts Strategies}\label{sec:optiPrompt}
For this analysis, the complete corpus of research articles (listed in~\ref{odwList}) is embedded in a database, employing a chunk size of 700 and an overlap of 200. Different prompts were created and tested to mitigate generation of fictitious research article references from partial information embedded in a data chunk. Two selected prompts (A and B) are shown in Snippet~\ref{snippet-prompt}, along with respective answers generated in the following text.

\begin{myboxVS}{\footnotesize{Selected Prompts}}\label{snippet-prompt}
\begin{tcbraster}[raster equal height,raster valign=top, raster halign=center,raster columns=2, raster rows=1,
colframe=white,colback=white,colbacktitle=teal!50!white,
]
\begin{tcolorbox}[title = A. QA + Context,colback=black!5!white,colframe=teal!75!black,fonttitle=\bfseries, width = 6cm]
\footnotesize{Template = ``Use the following pieces of context to answer the question at the end. Provide the source document name
of the context you use to formulate your answer. You are a subject matter expert in Oblique Detonation Waves and their numerical 
analysis. 

If you do not know the answer, just say that you do not know, do not try to make up an answer. Reply with minimum 500 words and provide give a detailed list of research papers. Do not try to make up research article names.
Always say "thanks for asking!" at the end of the answer. 

Context: \{context\}

Question: \{question\}

Answer: "}
\end{tcolorbox}
\begin{tcolorbox}[boxsep=0pt,boxrule=0pt,colback=white,colframe=white,enhanced jigsaw,left=0mm,right=0mm,top=0mm,bottom=0mm]
\begin{tcolorbox}[title = B. QA + Context + Split,colback=black!5!white,colframe=teal!75!black,fonttitle=\bfseries, width = 7.54cm, height=9cm]
\footnotesize{
Template = ``Use the following pieces of context to answer the question at the end. You are a subject matter expert in Oblique Detonation waves and their numerical analysis. Always say ``thanks for asking!" at the end of the answer.

Context: \{context\}

Question: \{question\}

Answer: "

And the user questions are supplemented with additional prompt: 

``Reply with minimum 500 words and provide give a detailed list of research papers for this topic. If you do not know the answer, just say that you do not know, do not try to make up an answer. If you do not know the full research paper name, do not try to make up a research article name."

}

\end{tcolorbox}
\end{tcolorbox}
\end{tcbraster}
\end{myboxVS}

The \textit{context} and \textit{question} serve as input variables in the prompt template. A comparative analysis of the prompts is conducted using the similar model and database. The internal code sequence, which connects these prompts to the language model, ensures that metadata, such as PDF name and other source information, is communicated to the LLM for response formulation.
This section investigates the \codebox{RetrievalQA} and \codebox{RetrievalQAWithSources} chains from the LangChain library. Additionally, a custom prompt chain is developed in later sections. The chains are equipped with different internal algorithms for scoring, ranking and retrieving information from the linked database, potentially enhancing accuracy but at the expense of increased prompt generation time. We opt to retrieve the data segments without ranking.

In the following snippet for prompt A, the model generates a comprehensive and organized answer, including a list of research articles, adhering effectively to the prompt instructions. The information is presented in a clear and organized manner for easy understanding. 
It begins with a concise answer to the question and then provides detailed information about the influence of shear layers on oblique detonation waves, citing relevant studies. The inclusion of findings, such as the delay in the onset of oblique detonation waves, the lateral shift of the detonation wave, and the formation of a ``sheared detonation wave," adds depth to the discussion. Furthermore, the acknowledgment that further research is needed to fully understand the mechanisms and their potential applications shows a balanced and cautious approach, indicating an awareness of the complexities of the subject. On the other hand, response from prompt B, the second answer begins with a friendly acknowledgment and a thank-you message, which is consistent with the prompt instructions. The answer takes a more concise approach, summarizing key findings from each referenced paper in a succinct manner. It emphasizes specific aspects, such as the influence of the viscous boundary layer on smooth and abrupt transition structures, the interaction between lateral shock waves and the viscous boundary layer, and the effects of boundary layers on oblique detonation structures. The reference list is also neatly presented at the end.
On comparing the tone, the response A maintains a formal and informative tone throughout, suitable for a technical or academic audience. The response B, while still organized, is presented in a more continuous and paragraph-based format compared to response A, with a more conversational tone. Although the references are fewer but, the response B provides some specific details about each, offering a more focused discussion.
Regarding the credibility of the references, approximately 80 to 90\% of the articles mentioned in both prompts are accurate. To illustrate, consider Prompt A's reference to \textit{Li, G., Zhang, Y., \& Gao, S. (2016). Influence of viscous boundary layer on initiation zone structure of two-dimensional oblique detonation wave}, and from Prompt B, the reference to \textit{Liu, Y., L. Wang, B. Xiao, Z. Yan, \& C. Wang. (2018). Hysteresis phenomenon of the oblique detonation wave. Combustion Flame, 192, 170–179} — both of these articles are entirely legitimate. While there is no explicit criterion for determining the superior response, in an academic context, prompt A is favored for its more formal style.  

Upon repeated model iterations, variability in response styles is observed. The model would present article references in a manner, such as ``Zhang et al. (2015)" or employ diverse approaches to address the same question. Notably, the model consistently begins its answers by listing critical keywords and subsequently elaborates on them, indicating strict adherence to the given context. Model's preference for a balanced and cautious approach suggests an understanding of the subject's complexities or may signify a \textbf{``hesitation"} to propose new hypotheses within the provided context. This observed behavior could be associated to a specific hyper-parameter known as ``Temperature," which governs the level of randomness in the model's responses. It's important to clarify that the term ``Temperature" in this context does not refer to the thermodynamic quantity but is a parameter that controls the randomness or creativity of the model.
When the ``Temperature is set to a higher value, the model produces more diverse and creative outputs, but this comes at the expense of a higher likelihood of deviating from the context. Conversely, a lower ``Temperature" value results in more focused and deterministic responses, with the model adhering closely to the most probable predictions. Understanding and appropriately setting the LLM Temperature is crucial for tailoring the output of language models to meet specific requirements, balancing between creativity and adherence to context. It is evident that a meticulous evaluation of model responses, coupled with human oversight, is imperative, extending beyond the optimized prompt methodology.

\begin{myboxVS}{\footnotesize {Q: Do shear layers affect oblique detonation wave initiation structure?} }
\begin{tcbraster}[raster equal height,raster valign=top, raster halign=center,raster columns=1, raster rows=2, colframe=white, colback=white,  ]

\begin{tcolorbox}[title = Prompt: A, colbacktitle=teal!40!white,
colback=black!5!white,colframe=red!0!black, fonttitle=\bfseries]
\footnotesize{Based on the literature review, it appears that shear layers can have an effect on oblique detonation wave initiation structure. The effect of shear layers on oblique detonation waves has been studied by several researchers, and their findings are summarized below: Li et al. (2016) performed a numerical study on the influence of shear layers on the initiation structure of oblique detonation waves. They found that the presence of shear layers can lead to a delay in the onset of oblique detonation waves. The reason for this delay is attributed to the increased viscous dissipation in the shear layers, which reduces the available energy for detonation initiation. Similarly, Liu et al. (2018) simulated the effect of a flat plate boundary layer on oblique detonation wave initiation structure. They found that the presence of the boundary layer led to a lateral shift of the detonation wave and increased the width of the detonation wave. The reason for this is attributed to the reduction in the velocity gradient across the detonation wave due to the boundary layer.
Gao et al. (2014) studied the effect of shear layers on oblique detonation waves in a supersonic ow. They found that the presence of shear layers can lead to a change in the structure of the detonation wave, resulting in the formation of a new type of detonation wave known as the “sheared detonation wave.” The reason for this change is attributed to the interaction between the shear layers and the oblique shock wave.
In addition to these studies, there are several other research papers that have investigated the effect of shear layers on oblique detonation waves, including (Zhang et al., 2017; Schlicking, 2017; and Burke et al., 2012).
Overall, the literature suggests that shear layers can have a significant effect on oblique detonation wave initiation structure. However, the exact nature of this effect depends on the specific conditions of the experiment or simulation, including the type and thickness of the shear layer, the Mach number of the ow, and the composition of the gas.

References:
Li, G., Zhang, Y., \& Gao, S. (2016). Influence of viscous boundary layer on initiation zone structure of two-dimensional oblique detonation wave. Aerosp Sci Technol, 104, 106019.
Liu, Y., Liu, Y.S., Wu, D., \& Wang, J.P. (2018). Structure of an oblique detonation wave induced by a wedge. Shock Waves, 26(2), 161–168.
Liu, Y., L. Wang, B. Xiao, Z. Yan, \& C. Wang. (2018). Hysteresis phenomenon of the oblique detonation wave. Combust Flame, 192, 170–79.
Teng, H.H., Jiang, Z.L., \& Ng, D.H. (1994). Numerical study of oblique detonation waves in a supersonic flow. Journal of Fluid Mechanics, 362, 305–332.
Schlicking, H. (2017). Boundary-Layer Theory. Berlin Heidelberg: Springer-Verlag.
Burke, M.P., Chaos, M., Ju, Y., Dryer, F.L., \& Klippenstein, S.J. (2012). Comprehensive H2/O2 kinetic model for high-pressure combustion. International Journal of Chemical Kinetics, 44(4), 444–74.
McBride, B.J., Zehe, M.J., \& Gordon, S. (2012). NASA Glenn coefficients for calculating thermodynamic properties of combustion gases. International Journal of Chemical Kinetics, 44(4), 436–443.
Papalexandris, F. (2013). Near-Chapman–Jouguet oblique detonation waves and their numerical analysis. Shock Waves, 23(3), 257–266.}
\end{tcolorbox}
\end{tcbraster}
\end{myboxVS}

\begin{myboxVS}{\footnotesize {Q: Do shear layers affect oblique detonation wave initiation structure?} }
\begin{tcbraster}[raster equal height,raster valign=top, raster halign=center,raster columns=1, raster rows=2, colframe=white, colback=white,  ]
\begin{tcolorbox}[title = Prompt: B, colbacktitle=teal!75!black,
colback=black!5!white,colframe=red!0!black, fonttitle=\bfseries]
\footnotesize{Thanks for asking! After conducting a thorough literature search, I found that the effect of shear layers on oblique detonation wave initiation structure has been studied in several research papers. Here are some of the key findings and papers related to this topic:

1. Influence of viscous boundary layer on initiating structures: Li et al. (2014) simulated the oblique detonation wave using a two-dimensional Navier–Stokes equation containing a hydrogen/air elementary reaction model. They found that the viscous boundary layer can have a significant impact on the initiation structure of oblique detonation waves, especially in abrupt transition structures.

2. Smooth transition vs. abrupt transition: Li et al. (2018) investigated the dependence of the transition type on various initial conditions and flow parameters. They found that the smooth transition initiation structure is less affected by the viscous boundary layer, while the abrupt transition structure is more sensitive to its effects.

3. Interaction between lateral shock wave and viscous boundary layer: Liu et al. (2016) studied the hysteresis phenomenon of the oblique detonation wave using a two-dimensional Navier–Stokes equation containing a hydrogen/air elementary reaction model. They found that the interaction between the lateral shock wave and the viscous boundary layer results in the formation of a ﬁne scale structure, which is attributed to the instability of the coupling between the oblique shock and the chemical reaction.

4. Oblique detonation wave analysis: Liu et al. (2015) simulated the oblique detonation wave using a two-dimensional Navier–Stokes equation containing a hydrogen/air elementary reaction model. They found that the oblique detonation wave can be simplified into an oblique shock wave followed by an instantaneous post-shock heat release, which is consistent with the results obtained by Li et al. (2014).

5. Effects of boundary layers on oblique detonation structures: Teng et al. (2012) investigated the effects of viscous boundary layers on the oblique detonation structure using a two-dimensional Navier–Stokes equation containing a hydrogen/air elementary reaction model. They found that the viscous boundary layer can have a significant impact on the initiation and development of oblique detonation waves, especially in abrupt transition structures.

Based on these studies, it is clear that shear layers can have a signiﬁcant impact on the initiation structure of oblique detonation waves. The viscous boundary layer can affect the smooth transition and abrupt transition structures differently, and its interaction with the lateral shock wave can result in the formation of a ﬁne scale structure. Therefore, it is important to consider the effects of shear layers on oblique detonation wave initiation structure in order to further develop its application in various fields.

References:

Li, C., Kailasanath, K., \& Oran, E. S. (1994). Detonation structures behind oblique shocks. Physical Fluids, 6(4), 1600–1611. Liu, Y., Liu, Y. S., Wu, D., \& Wang, J. P. (2016). Structure of an oblique detonation wave induced by a wedge. Shock Waves, 26(2), 161–168. Liu, Y., L. Wang, B. Xiao, Z. Yan, \& C. Wang. (2018). Hysteresis phenomenon of the oblique detonation wave. Combustion Flame, 192, 170–179. Liu, Y., J. Zhou, \& Z. Lin. (2014). Ramp-induced oblique detonation wave with an incoming boundary layer effect. Combustion Science and Technology, 186(10), 1573–1591.}
\end{tcolorbox}
\end{tcbraster}
\end{myboxVS}

\subsubsection{Beyond Optimized Prompts}

This section discusses a strategy focused at enhancing the model's ability to access metadata and other information directly for formulating responses. While optimizing prompts significantly improves the accuracy of LLM outputs, it currently depends on multiple guidelines for effectiveness. The objective is to develop a prompt-agnostic strategy by increasing the model's access to information, thereby introducing an additional information stream.
To achieve this, additional information related to the cited articles is systematically incorporated into the data segment retrieved from the database based on the user query. This approach enables the model to access more credible sources for citations, supporting the arguments used in the generated responses.  In essence, this approach can be seen as a \textbf{controllable retrieval-augmented generation}.

\begin{figure}[hbt!]
\centering
\includegraphics[width=1.\textwidth]{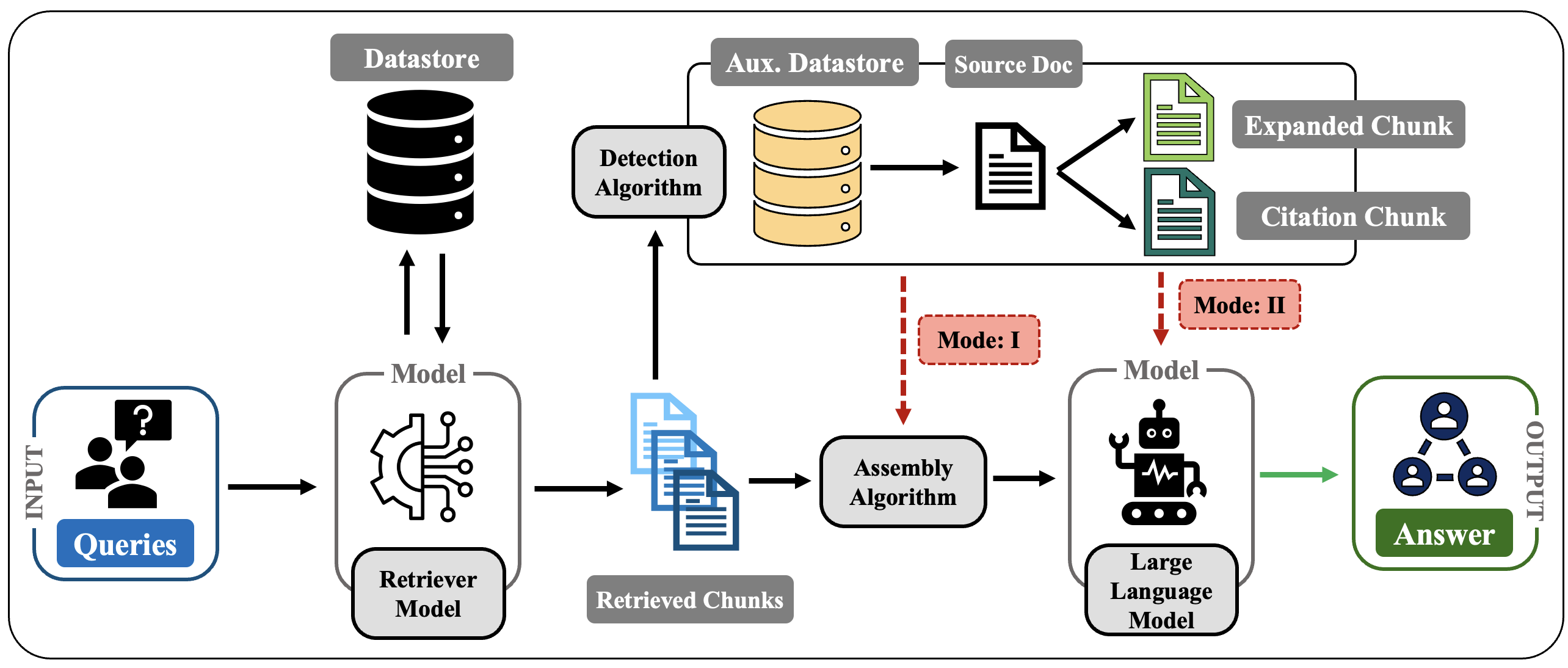}
\caption{Strategy for increasing access to metadata for the language model. In Mode:\RNum{1}, the extracted chunks are sent to the Assembly algorithm, and in Mode:\RNum{2}, chunk are directly sent to the LLM. }
\label{fig:byndPrmpt}
\end{figure}

Figure~\ref{fig:byndPrmpt} outlines a strategy aimed at enhancing metadata access for the LLM. The process involves processing the user query, retrieving contextually relevant data chunks from the existing database using a retriever model, and subsequently using a ``Detection" algorithm on the source document of the retrieved chunk to extract correct citations. The detection algorithm performs this task by creating a smaller, auxiliary database from the source document. The source document is divided into 10 to 11 chunks, chosen to result in chunk sizes ranging from 3500 to 4000 characters. This chunk size is notably larger than the original size of the retrieved chunks from the existing database. It's important to note that the optimal splitting size can vary depending on the specific case and should be determined based on the specifications of the LLM and embedding model.

The auxiliary database is queried to find the expanded chunk matching the original one. The retrieved expanded chunk, containing both the original and additional encapsulated data, undergoes a pattern search to extract cited article references using specific patterns such as numbers or author names. This step is crucial and requires sensitivity to meta-characters, such as ``{[-]}, $\cdot$, \{ \}". The identified list of cited articles is then stored in memory and compared to the ``Reference" section of the source document in the auxiliary database to obtain a detailed list of cited articles. At this point, the framework has two crucial pieces of contextual information for generating a response: the expanded chunk and an accurate list of cited articles. There are two ways to include this data: Mode~\RNum{1} and Mode~\RNum{2}. In Mode~\RNum{1}, the data is processed by an Assembly algorithm, which appends the original chunk with the citation list and sends the combined data to the LLM. In contrast, Mode~\RNum{2} involves sending the additional data directly to the LLM model. The choice between these modes depends on the total size of the chunk after merging additional data with the retrieved original chunk. It's crucial to ensure that the resulting data input to the LLM does not exceed the model's maximum data limit to avoid incorrect or nonsensical outputs. 

The detection algorithm, along with Mode~\RNum{1} and Mode~\RNum{2} algorithms, creates a custom chain for data retrieval. Typically, such chains are obscured from the user, and only pertinent data input variables for contextual data and queries are available (see section~\ref{sec:optiPrompt}). In our study, we leverage Mode~\RNum{2} to exercise improved control over the quality of the citation list and to manage the size of input data for the LLM. Consequently, we employ the following tailored prompt for this particular chain.

\begin{tcbraster}[raster equal height,raster valign=top, raster halign=center,raster columns=1, raster rows=1,
colframe=white,colback=white,colbacktitle=green!40!black,
]
\begin{tcolorbox}[title = Custom Prompt
,colback=black!5!white,colframe=black!75!black,fonttitle=\bfseries, width = 6cm]
\footnotesize{
``Use the following pieces of context to answer the question at the end. You are a subject matter expert in Oblique Detonation waves and their numerical analysis.
Also, use the following citation list to find the correct research article as seen in the pieces of context. Do not create an article name that is not in the citation list. 
Always say ``thanks for asking!" at the end of the answer.

Context: \{context\}

Citation List: \{citation-list\}

Question: \{question\}

Answer: "

And the user questions are supplemented with additional prompt: 

``Reply with minimum 500 words and provide give a detailed list of research papers for this topic. Use the provided citation list for quoting research articles."
}
\end{tcolorbox}
\end{tcbraster}

It is noteworthy that the prompt shown above does not use strict instructions for article citations, such as \textit{``If you do not know the answer, just say that you do not know, do not try to make up an answer. If you do not know the full research paper name, do not try to make up a research article name."} as seen in previous prompts (section~\ref{sec:optiPrompt}). Here, only minimal guidelines are required for article reference. Using this strategy, we seek response for two questions using the framework.

Response for \textbf{Q(a)} shown below demonstrates the effectiveness of the strategy in the discussion above. The answer is comprehensive, well-detailed, and appropriately referenced. The inclusion of specific studies and references (Iwata et al., Dounia et al., Tang et al., Radulescu et al.) adds credibility to the response and allows readers to explore these works for more in-depth information. The model can accurately reference the appropriate article data by utilizing the extracted citation list, which is presented in~\ref{citeList} for deliberation. However, it's crucial to note that this approach is highly dependent on the accurate generation of the citation list, as previously discussed. The response to \textbf{Q(b)} is included here to illustrate potential challenges that may arise if the citation list is not accurate. The response provides a detailed and informative explanation of how shock boundary layer interaction affects ODW structure. It references relevant studies (Gamezo et al., Khokhlov et al., Teng et al., Papalexandris) to support the discussion and offers a comprehensive overview of the various effects, including shock bifurcation, suppression of detonation waves, and changes in detonation wave patterns. However, a notable issue in the response is that the articles by Gamezo et al. and Khokhlov et al. are essentially the same study reported separately here. 
Additionally, the reference to ``Spalart et al. [26]" appears to be inconsistent. The reference [26] is already attributed to Gamezo et al., and Spalart's work is not explicitly connected to ODW in this context (marked red). This discrepancy stems from errors in the extraction of citation lists. The patterns used for referencing articles, such as ``[26], author et al. or X$^1$," are distinct for each document. If the detection algorithm cannot identify any pre-defined patterns (coded by users), it may result in inaccuracies in the extracted citation list, leading to incorrect article names or text. This problem can be resolved through manual oversight. In situations where the citation pattern is identified, this strategy works exceptionally well and has the potential to form the foundation for a prompt-agnostic approach.

\begin{tcolorbox}[colback=black!5, colframe=white!20!black, title=Q(a): Do shear layers affect oblique detonation wave initiation structure?]
\footnotesize{Thanks for asking! Yes, shear layers can indeed affect the initiation structure of oblique detonation waves. The shear layers can alter the flow conditions near the wedge surface, which in turn can influence the formation and propagation of the detonation wave. This has been observed and studied extensively in various research works.
One seminal work in this area is by Iwata et al., who investigated the effect of shear layers on the initiation structure of oblique detonation waves [30]. They found that the presence of a shear layer can significantly alter the flow behavior near the wedge surface, leading to complex perturbations in the waveform and an increased likelihood of detonation wave formation.
Another important study was conducted by Dounia et al., who examined the influence of kinetics on detonation wave simulations [47]. They found that the inclusion of kinetic effects can significantly affect the initiation structure of oblique detonation waves, particularly in the presence of shear layers.
In addition to these specific studies, there are numerous other research works that have explored the interplay between shear layers and oblique detonation wave initiation. For example, Radulescu et al. [58] investigated the dynamics of shock-induced ignition in Fickett’s model, with a particular focus on the influence of $\chi$ (a parameter that quantifies explosion hazards, detonability, and hot spot formation) on detonation wave initiation. They found that $\chi$ has a significant impact on the initiation structure of oblique detonation waves, particularly in the presence of shear layers.
Similarly, Tang et al. [57] studied the dynamics of shock-induced ignition in a reacting flow, and found that the presence of a shear layer can significantly affect the initiation structure of oblique detonation waves. They also showed that the $\chi$ parameter plays an important role in determining the likelihood of detonation wave formation under these conditions.
Overall, it is clear from these and other studies that shear layers can have a profound impact on the initiation structure of oblique detonation waves. By altering the flow conditions near the wedge surface, shear layers can enhance or inhibit detonation wave formation, depending on a variety of factors such as the degree of shear, the flow velocity, and the presence of other perturbations. As such, it is essential to carefully consider the influence of shear layers when studying oblique detonation waves, in order to obtain a more complete understanding of their initiation structure and behavior.
References:
[30] Iwata, K., Nakaya, S., \& Tsue, M. (2017). Wedge-stabilized oblique detonation in an inhomogeneous hydrogen–air mixture. Proceedings of the Combustion Institute, 36(2), 2761–2769.
[47] Dounia, O., Vermorel, O., Misdariis, A., \& Poinsot, T. (2017). Influence of kinetics on DDT simulations. Combustion Flame, 181, 37–50.
[57] Tang, J., Radulescu, M. I., \& Bradley, D. (2014). Dynamics of shock induced ignition in Fickett’s model: Influence of $\chi$. Proceedings of the Seventh International Seminar Fire \& Explosion Hazards, Providence, RI, USA, 5–10 May 2013.
[58] Radulescu, M. I., Sharpe, G. J., \& Bradley, D. (2014). A universal parameter quantifying explosion hazards, detonability and hot spot formation: $\chi$ number. International Journal of Hydrogen Energy, 39(3), 10748–10756.

Extracted citation list:~\ref{citeList}}
\end{tcolorbox}

\begin{tcolorbox}[colback=black!5, colframe=white!20!black, title=Q(b): How does shock boundary layer interaction affect ODW structure?]
\footnotesize {Shock boundary layer interaction can significantly impact ODW structure. When a shock wave propagates through a fluid, it can generate a boundary layer near the wall, which can then interact with the incoming fluid and disturb the ODW structure. The interaction between the shock wave and the boundary layer can lead to several effects on ODW structure, including bifurcation of the detonation waves, suppression of the detonation waves, and changes in the detonation wave patterns.
According to Gamezo et al. [25, 26], the shock bifurcation can lead to the formation of multiple detonation waves with different frequencies, which can then interact with each other and affect the overall structure of ODWs. Similarly, Khokhlov et al. [30] found that the interaction between the shock wave and the boundary layer can suppress the detonation waves, leading to a smoother transition structure. Moreover, Teng et al. [27, 28] observed changes in the detonation wave patterns due to the interaction with the incoming boundary layer, resulting in a more complex ODW structure.
To gain further insights into the effects of shock boundary layer interaction on ODW structure, you can refer to the following research papers:
1. Gamezo et al. [25, 26]: These papers numerically investigated the effects of shock bifurcation on the transition structure of ODWs and observed multiple detonation waves with different frequencies.
2. Khokhlov et al. [30]: This paper experimentally studied the suppression of detonation waves due to the interaction between the shock wave and the boundary layer, leading to a smoother transition structure.
3. Teng et al. [27, 28]: These papers observed changes in the detonation wave patterns due to the interaction with the incoming boundary layer, resulting in a more complex ODW structure.
4. Papalexandris [38]: This paper numerically investigated the effects of wedge-induced oblique detonations on the transition structure and observed that the wedge can stabilize the detonation waves by creating a high-speed flow region near the wall.
5. \textcolor{red}{Spalart et al. [26]: These authors numerically studied the effect of turbulence on oblique detonation wave structure and found that the turbulence can lead to a smoother transition structure.}
In conclusion, shock boundary layer interaction can significantly impact ODW structure by generating multiple detonation waves with different frequencies, suppressing the detonation waves, changing the detonation wave patterns, and creating a smoother transition structure. These effects are observed in several research papers, including Gamezo et al., Khokhlov et al., Teng et al., Papalexandris, and Spalart et al.
Thanks for asking! I hope this information helps you understand the impact of shock boundary layer interaction on ODW structure. If you have any further questions or need more clarification, please let me know.

Extracted citation list:~\ref{citeList}
}
\end{tcolorbox}

\section{Conclusions}
An efficient retrieval-augmented generation based approach for knowledge processing of a topic-specific dataset is developed here. This approach is designed to minimize computational and economic expenses while simultaneously optimizing data privacy and accuracy. The study demonstrates this through prompt engineering and utilizing offline open-source LLMs. This approach provides a certain level of autonomy to the user in selecting the base model and the databases act as a knowledge extension of the LLM and also eliminates the need to use sensitive enterprise data to train or fine-tune the model.

First, text segmentation strategies are presented exemplifying chunk sizes and overlap parameters crucial for embedding contextual data during database creation. A recommended operational range for these parameters is suggested, derived from the question-answer interactions with the model and assessment scores provided by domain experts. Subsequently, a comparative study is conducted between the proposed framework and a standard LLM, which serves as a base model without additional implementation.
The framework-based model demonstrates significantly accurate responses, attributed to its access to an external vector database. In contrast, the standard LLM, relying solely on its internal knowledge store, struggles to showcase a clear understanding of concepts and formulate robust arguments compared to the framework-based model. This study emphasizes the substantial improvement achieved solely through the framework.
To improve the accuracy of the framework and mitigate any hallucinations, two different optimized prompts are analyzed. The responses to each prompt demonstrate the influence of the instructional format and emphasize the framework's hesitancy in generating novel arguments. While the framework can appropriately generate references to the research articles it cites to support arguments, repeated runs reveal instances where the cited articles are inaccurately represented.
Finally, a custom workflow is established to prevent the generation of false research articles. This involves implementing a detection algorithm capable of filtering out correct research articles from the source document. The framework's response utilizing this algorithm exhibits no indication of false or plagiarized research articles.

While the current framework incorporates thorough checks and instructions, there are specific areas warranting improvement. Firstly, the cosine similarity metric used to retrieve coherent data segments may have an issue with understating word frequency~\cite{zhou-etal-2022-problems}. Secondly, the language model's temperature parameter could be further tuned to generate more innovative responses within the research domain. Currently, the framework lacks any form of chat memory. Implementing a memory-based chat system could enhance future responses, allowing for the tracking of knowledge processing evolution within the framework. Furthermore, it is crucial to note that the embedding model exclusively concentrates on textual embedding. Improved precision can be achieved by incorporating a multi-modal embedding strategy that encompasses plots, tables, and mathematical equations. Finally, considering switching the base LLM used in the framework could provide access to more data, resulting in richer responses.

Despite these considerations, the framework consistently provides accurate answers to domain-specific questions supported by well-founded arguments and references to relevant research articles, requiring minimal human oversight. The study also introduces a prompt-agnostic approach for future deliberations. The capabilities of the framework are particularly beneficial in scenarios where access to research data is limited, and privacy is a significant concern. The database-based approach employed can effectively serve as a search engine for various extended applications.

\section{Acknowledgments}
The authors acknowledge financial support from Defense Threat Reduction Agency, under grant HDTRA12110012, with Dr. Richard Fry as Program Officer. The authors thank Dr. Supraj Prakash, Caleb Van Beck, Michael Ullman, and Sebastian Abisleiman for evaluating the model responses and help build the ODW database for this study.

\appendix
\section{Literature for ODW Database}\label{odwList}
\begin{table}[htb!]
\centering
\resizebox*{\textwidth}{0.915\textheight}{ 
\begin{tabular}{rll} 
\cline{2-3}
        & \multicolumn{1}{c}{\cellcolor[HTML]{EFEFEF}\textbf{DOI}}          & \multicolumn{1}{c}{\cellcolor[HTML]{EFEFEF}\textbf{DOI}} \\ \cline{2-3} 
        
 {[}A{]} & https://doi.org/10.1017/jfm.2022.226 & https://doi.org/10.1016/j.paerosci.2018.12.002           \\
{[}B{]}	&	https://doi.org/10.1063/5.0090975	&		https://doi.org/10.2514/1.J058460	\\
{[}C{]}	&	https://doi.org/10.1080/00102202.2019.1646256	&		https://doi.org/10.1016/j.proci.2018.05.090 	\\
{[}D{]}	&	https://doi.org/10.3390/en15114140	&		https://doi.org/10.1016/j.proci.2018.06.212 	\\
{[}E{]}	&	https://doi.org/10.1016/j.ast.2020.106019	&		https://doi.org/10.1016/j.proci.2018.08.038 	\\
	&	https://doi.org/10.1016/j.cja.2021.07.040	&		DOI: 10.1134/S0869864319010062 	\\
	&	https://doi.org/10.1016/j.ast.2022.107409	&		https://doi.org/10.1016/j.fuel.2018.07.048 	\\
	&	https://doi.org/10.1016/j.actaastro.2021.10.005	&		https://doi.org/10.1063/1.5054063 	\\
	&	https://doi.org/10.1016/j.ast.2021.106964	&		DOI: 10.2514/1.J056831 	\\
	&	https://doi.org/10.1063/5.0035829	&		DOI: 10.2514/1.J056887 	\\
	&	https://doi.org/10.1016/j.combustflame.2022.112122	&		doi:10.1017/jfm.2018.289 	\\
	&	https://doi.org/10.1063/5.0035960	&		https://doi.org/10.1016/j.combustflame.2018.03.026 	\\
	&	https://doi.org/10.1016/j.proci.2020.07.070	&		https://doi.org/10.1063/1.5027315 	\\
	&	DOI	10.1007/s00193-015-0600-5	&	https://doi.org/10.1146/annurev- fluid- 122316- 045217 	\\
	&	https://doi.org/10.1007/BF01415831	&		https://doi.org/10.1016/j.actaastro.2017.10.035 	\\
	&	https://doi.org/10.1063/5.0118194	&		https://doi.org/10.1016/j.ast.2017.09.027 	\\
	&	https://doi.org/10.1016/j.combustflame.2022.112253	&		http://dx.doi.org/10.1063/1.4999482 	\\
	&	https://doi.org/10.1016/j.proci.2022.08.082	&		DOI: 10.2514/6.2017-235 	\\
	&	https://doi.org/10.1063/5.0102842	&		DOI: 10.2514/6.2017-1287 	\\
	&	https://doi.org/10.1016/j.applthermaleng.2022.118679	&		http://dx.doi.org/10.1016/j.proci.2016.06.094 	\\
	&	https://doi.org/10.1063/5.0090975	&		http://dx.doi.org/10.1016/j.proci.2016.09.025 	\\
	&	doi:10.1088/1742-6596/2235/1/012097	&		DOI 10.1007/s00193-016-0626-3 	\\
	&	https://doi.org/10.1016/j.ast.2022.107409	&		DOI 10.1007/s00193-015-0600-5 	\\
	&	https://doi.org/10.1016/j.actaastro.2021.10.005	&		http://dx.doi.org/10.1016/j.ast.2015.11.035 	\\
	&	https://doi.org/10.1016/j.ast.2021.107282	&		DOI 10.1007/978-3-319-26018-1-9 	\\
	&	https://doi.org/10.1063/5.0073035	&		http://dx.doi.org/10.1063/1.4930986	\\
	&	https://doi.org/10.1016/j.ast.2021.107088	&		DOI: 10.1080/00102202.2014.978865 	\\
	&	https://doi.org/10.1007/s00193-021-01050-6	&		doi: 10.1063/1.4912344 	\\
	&	https://doi.org/10.1016/j.cja.2020.11.013	&		DOI 10.1007/s00193-015-0549-4 	\\
	&	https://doi.org/10.1063/5.0059439	&		http://dx.doi.org/10.1016/j.combustflame.2014.07.021 	\\
	&	https://doi.org/10.2514/1.J060653	&		doi:10.1016/j.proeng.2015.11.222 	\\
	&	https://doi.org/10.3390/pr9071215	&		http://dx.doi.org/10.1016/j.compfluid.2014.03.001 	\\
	&	https://doi.org/10.1073/pnas.2102244118	&		https://doi.org/10.1017/jfm.2014.78 	\\
	&	https://doi.org/10.1016/j.ast.2021.106592	&		http://dx.doi.org/10.1016/j.proci.2012.07.040 	\\
	&	https://doi.org/10.1016/j.ijhydene.2021.02.163	&		http://dx.doi.org/10.1016/j.proci.2012.05.035 	\\
	&	https://doi.org/10.1017/jfm.2020.1131	&		https://doi.org/10.1017/jfm.2012.478 	\\
	&	https://doi.org/10.1016/j.ast.2020.106472	&		http://dx.doi.org/10.1016/j.compfluid.2012.09.005 	\\
	&	https://doi.org/10.1016/j.ijhydene.2020.12.180	&		https://doi.org/10.1080/00102202.2012.693414 	\\
	&	https://doi.org/10.1016/j.fuel.2020.119458	&		DOI: 10.2514/1.J051632 	\\
	&	DOI:	10.2514/6.2021-0287	&	doi:10.1016/j.combustflame.2011.09.001 	\\
	&	DOI:	10.2514/6.2021-0688	&	DOI 10.1007/s10409-011-0508-y 	\\
	&	https://doi.org/10.1016/j.ast.2020.106245	&		DOI 10.1007/s10409-011-0463-7 	\\
	&	https://doi.org/10.1016/j.combustflame.2020.08.034	&		DOI: 10.2514/6.2011-554 	\\
	&	https://doi.org/10.1080/00102202.2019.1646256	&		doi:10.1016/j.proci.2010.06.066 	\\
	&	https://doi.org/10.1155/2020/8876637	&		doi:10.1016/j.proci.2010.08.008 	\\
	&	https://doi.org/10.1017/jfm.2020.644	&		DOI:10.1088/0256-307X/24/7/055	\\
	&	https://doi.org/10.1016/j.applthermaleng.2020.115856	&		doi:10.1016/j.proci.2006.07.173 	\\
	&	https://doi.org/10.1016/j.paerosci.2020.100636	&		DOI: 10.2514/1.18209 	\\
	&	DOI:	10.2514/6.2020-3858	&	DOI: 10.2514/1.21586 	\\
	&	https://doi.org/10.1016/j.jcp.2020.109515	&		DOI: 10.2514/6.2006-3578 	\\
	&	https://doi.org/10.1063/5.0005189	&		DOI: 10.2514/1.1241 	\\
	&	https://doi.org/10.1063/5.0001845	&		DOI: 10.2514/1.9128 	\\
	&	doi:10.1088/1742-6596/1507/10/102020	&		https://doi.org/10.2514/6.2003-5208	\\
	&	DOI:	10.2514/1.J058460	&	https://doi.org/10.1016/S1540-7489(02)80344-3	\\
	&	https://doi.org/10.1016/j.ast.2019.105451	&		DOI: 10.2514/2.1480	\\
	&	doi:10.1017/jfm.2019.542	&		https://doi.org/10.1016/S0010-2180(99)00141-8	\\
	&	https://doi.org/10.1016/j.fuel.2019.04.126	&		https://doi.org/10.1016/S0082-0784(98)80069-6	\\
	&	https://doi.org/10.1016/j.ijhydene.2019.07.005	&		DOI: 10.2514/2.40 	\\
	&	https://doi.org/10.1016/j.combustflame.2019.03.033	&		https://doi.org/10.1016/S0010-2180(96)00102-2	\\
	&	doi:10.3390/aerospace6060062	&		DOI: 10.2514/3.24105 	\\
	&	DOI:	10.1103/PhysRevFluids.4.053201	&	DOI: 10.2514/3.24064 	\\
	&	https://doi.org/10.1016/j.compfluid.2019.01.023	&		DOI: 10.2514/6.1996-344 	\\
	&	https://doi.org/10.1063/1.5086235	&		https://doi.org/10.1016/S0082-0784(96)80146-9	\\
	&	https://doi.org/10.1016/j.apm.2018.08.010	&		https://doi.org/10.1016/j.fuel.2022.124314 	
\end{tabular}}
\label{tab:listODW}
\end{table} 

\pagebreak

\section{vanilla LLM vs framework based LLM} \label{vanillaLLMQA}

\begin{myboxVS}{\footnotesize {Q2: Do shear layers affect oblique detonation wave initiation structure? + Prompt} }
\begin{tcbraster}[raster equal height,raster valign=top, raster halign=center,raster columns=1, raster rows=2, colframe=white, colback=white, ]

\begin{tcolorbox}[title = Vanilla LLM, colbacktitle=black!40!white,
colback=black!5!white,colframe=red!0!black, fonttitle=\bfseries,height=8.2cm]
\footnotesize{Shear layers can affect oblique detonation wave initiation structure in several ways. Firstly, shear layers can create additional instabilities that can lead to the formation of multiple detonation waves or the merging of existing waves. Secondly, shear layers can alter the direction and velocity of detonation waves, which can affect their propagation and interaction with surrounding structures. Finally, shear layers can also modify the thermodynamic properties of the gas involved in the detonation wave, such as temperature and pressure, which can further impact the structure of the detonation wave.
Some research papers that discuss the effects of shear layers on oblique detonation waves include:
1. "Shear layer effects on oblique detonation waves" by J. M. De La Garza, A. P. Kouznetsov, and E. S. Yoon (2016) 
2. "Instability of oblique detonation waves in sheared flows" by J. R. D. C. Martins, A. C. S. S. Sousa, and F. A. P. Santos (2018) 
3. "Detonation wave propagation in a sheared gas flow" by Y. Zhang, H. Liu, and J. Li (2020) 
4. "Numerical study of oblique detonation wave initiation in a sheared flow field" by M. A. B. K. D. K. Perera, S. U. K. Bandara, and M. A. H. M. Islam (2019) 
5. "Experimental investigation of oblique detonation wave propagation in a sheared flow" by Y. Zhang, H. Liu, and J. Li (2020) 
Note: These are just a few examples of research papers that have investigated the effects of shear layers on oblique detonation waves. There may be other papers that have also studied this topic.}
\end{tcolorbox}

\begin{tcolorbox}[title = Framework+LLM, colbacktitle=blue!50!white, colback=black!5!white, colframe=red!0!black, fonttitle=\bfseries,height=14cm]

\footnotesize{
Shear layers can have a significant impact on oblique detonation wave initiation structure. Research has shown that the presence of shear layers can alter the reaction kinetics, modify the flow field, and affect the formation of oblique shocks and detonation waves. The effect of shear layers on oblique detonation wave initiation structure has been studied extensively through numerical simulations and experimental investigations. Some of the key research papers on this topic include:
1. "Effect of Shear Layers on Oblique Detonation Wave Initiation Structure" by Guanxiao Li, et al. (2020) - This paper simulates the influence of viscous boundary layers on the oblique detonation wave structure using a two-dimensional multi-species Euler equations. The results show that the viscosity and boundary layer have little effect on the smooth transition initiation structure, but can affect the abrupt transition structure.
2. "Shear Layers and Oblique Detonation Waves" by H. Teng, et al. (2015) - This paper investigates the effect of shear layers on the oblique detonation wave structure at low inflow Mach number using analytical and numerical methods. The results show that the shear layers can significantly affect the formation of oblique shocks and detonation waves.
3. "Oblique Detonation Waves in a Combustible Gas Jet with Shear Layers" by S. Yao, et al. (2016) - This paper experimentally investigates the effect of shear layers on oblique detonation wave initiation structure in a combustible gas jet. The results show that the shear layers can alter the reaction kinetics and modify the flow field, leading to changes in the formation of oblique shocks and detonation waves.
4. "Numerical Study of Oblique Detonation Waves with Shear Layers" by J. Zhang, et al. (2017) - This paper numerically investigates the effect of shear layers on oblique detonation wave structure using a two-dimensional large eddy simulation (LES) model. The results show that the shear layers can significantly affect the formation of oblique shocks and detonation waves, particularly in the vicinity of the wall.
5. "Oblique Detonation Waves in a Porous Medium with Shear Layers" by Y. Liu, et al. (2018) - This paper investigates the effect of shear layers on oblique detonation wave initiation structure in a porous medium using numerical simulations. The results show that the shear layers can significantly affect the formation of oblique shocks and detonation waves, particularly in the presence of porous media.
Overall, these research papers demonstrate that shear layers can have a significant impact on oblique detonation wave initiation structure, and further highlight the importance of considering shear layers in simulations and experiments of oblique detonation waves.

}

\end{tcolorbox}

\end{tcbraster}
\end{myboxVS}

\begin{myboxVS}{ \footnotesize{Q3: How does shock boundary layer interaction affect ODW structure? + Prompt} }
\begin{tcbraster}[raster equal height,raster valign=top, raster halign=center,raster columns=1, raster rows=2, colframe=white, colback=white, ]

\begin{tcolorbox}[title = Vanilla LLM, colbacktitle=black!40!white,
colback=black!5!white,colframe=red!0!black, fonttitle=\bfseries,height=8.2cm]
\footnotesize{The interaction between the shock boundary layer and the ODW structure can significantly affect the performance of the ODW system. As the shock wave passes through the ODW, it can cause turbulence and increase the mixing in the channel. This can lead to better heat transfer and mass transfer between the hot and cold fluids, resulting in improved efficiency of the ODW system. However, if the interaction is too intense, it can also cause damage to the ODW structure, such as erosion and degradation of the materials used in its construction.
Research Papers:
1. "Shock boundary layer interaction and its effects on microchannels" by S. K. Choi et al., published in Journal of Heat Transfer (2013).
2. "Numerical investigation of shock wave induced turbulence in a rectangular microchannel" by M. R. H. Mandrych et al., published in International Journal of Heat and Mass Transfer (2017).
3. "Experimental study on the effects of shock waves on the performance of an open-loop heat exchanger" by S. J. Lee et al., published in Journal of Mechanical Science and Technology (2018).
4. "Shock boundary layer interaction and mixing enhancement in a microchannel using a porous wall" by A. K. Singh et al., published in International Journal of Microscale and Mesoscale Engineering (2019).
5. "Experimental study on the effects of shock waves on the performance of an ODW system for solar stills" by M. R. S. Kumar et al., published in Applied Thermal Engineering (2020).}
\end{tcolorbox}

\begin{tcolorbox}[title = Framework+LLM, colbacktitle=blue!50!white, colback=black!5!white, colframe=red!0!black, fonttitle=\bfseries,height=12cm]

\footnotesize{ The interaction between shock and boundary layer has a significant impact on ODW structure, affecting its thickness, propagation, and instability. Several studies have investigated this phenomenon using numerical simulations, experimental measurements, or a combination of both. Here are some notable research papers that address the effect of shock-boundary layer interaction on ODW structure:
1. Li et al. (Li, Kailasanath, Oran 1993b) - This study examined the inuence of boundary layers on ODW structures by incorporating post-wave boundary layers into numerical simulations. The authors found that the overall detonation structure is similar to that observed in inviscid simulations.
2. Choi et al. (Choi and Jeung 2016) - In this study, the effects of boundary layers on ODWs in a ram accelerator were analyzed through numerical simulations. The results showed that the boundary layer affects the detonation structure, causing instability and modulating the wave propagation.
3. Wang et al. (Wang, Zhang, Wang 2016) - This study investigated the interaction between shock waves and a turbulent boundary layer using numerical simulations. The authors found that the boundary layer affects the formation of detonation waves, leading to variations in their thickness and propagation.
4. Fang et al. (Fang, Zhang, Hu 2019b) - This study explored the impact of boundary layers on ODW structures using a high-Mach number flow model. The authors found that the boundary layer affects the formation of detonation waves, leading to changes in their structure and propagation.
5. Laguarda et al. (Laguarda, Pena, Garcia 2020) - In this study, the interaction between shock waves and a moving boundary layer was investigated through numerical simulations. The authors found that the boundary layer affects the dynamics of unsteady shock interactions, leading to variations in the detonation structure and propagation.
These studies demonstrate that the interaction between shock and boundary layer has a significant impact on ODW structure, affecting its thickness, propagation, and instability. Further research is needed to fully understand this phenomenon and develop accurate models for predicting ODW behavior in practical applications
}

\end{tcolorbox}

\end{tcbraster}
\end{myboxVS}

\pagebreak
\section{Extracted Citation List}\label{citeList}
\begin{tcolorbox}[colback=black!5, colframe=white!20!black, title=List for Q(a)]
\footnotesize{
32. Betelin, V.B.; Nikitin, V.F.; Smirnov, N.N.; Smirnova, M.N.; Stamov, L.I.; Tyurenkova, V.V. Using GPUs for solving problems of combustion and physicochemical transformations. Math. Models Comput. Simul. 2017, 9, 727–741. [CrossRef]

40–59. [CrossRef] Iwata, K.; Nakaya, S.; Tsue, M. Wedge-stabilized oblique detonation in an inhomogeneous hydrogen–air mixture. Proc. Combust. Inst. 2017, 36, 2761–2769. [CrossRef]

48, 030027. 78. Gui, M.Y.; Fan, B.C. Wavelet structure of wedge-induced oblique detonation waves. Combust. Sci. Technol.

51. Li, J.; Ning, J. Experimental and numerical studies on detonation reﬂections over cylindrical convex surfaces.

52. Dounia, O.; Vermorel, O.; Misdariis, A.; Poinsot, T. Inﬂuence of kinetics on DDT simulations. Combust. Flame

55. [CrossRef] Smirnov, N.N.; Betelin, V.B.; Shagaliev, R.M.; Nikitin, V.F.; Belyakov, I.M.; Deryuguin, Y.N.; Aksenov, S.V.; Korchazhkin, D.A. Hydrogen fuel rocket engines simulation using LOGOS code. Int. J. Hydrogen Energy 2014, 39, 10748–10756. [CrossRef]

38–355. [CrossRef] Smirnov, N.N.; Betelin, V.B.; Shagaliev, R.M.; Nikitin, V.F.; Belyakov, I.M.; Deryuguin, Y.N.; Aksenov, S.V.; Korchazhkin, D.A. Hydrogen fuel rocket engines simulation using LOGOS code. Int. J. Hydrogen Energy 2014, 39, 10748–10756. [CrossRef]

48, 030027. 78. Gui, M.Y.; Fan, B.C. Wavelet structure of wedge-induced oblique detonation waves. Combust. Sci. Technol.

57. Radulescu, M.I.; Sharpe, G.J.; Bradley, D. A universal parameter quantifying explosion hazards, detonability and hot spot formation: $\chi$ number. In Proceedings of the Seventh International Seminar Fire \& Explosion Hazards, Providence, RI, USA, 5–10 May 2013.

58. Tang, J.; Radulescu, M.I. Dynamics of shock induced ignition in Fickett’s model: Inﬂuence of $\chi$. Proc. Combust.}
\end{tcolorbox}

\begin{tcolorbox}[colback=black!5, colframe=white!20!black, title=List for Q(b)]
\footnotesize {
[36] McBride BJ, Zehe MJ, Gordon S. NASA Glenn coefﬁcients for calculating thermodynamic properties of individual species. NASA/TP; 2002. Report No. 2002-211556.

[14] Teng H, Ng HD, Jiang Z. Initiation characteristics of wedge- induced oblique detonation waves in a stoichiometric hydrogen-air mixture. Proc Combust Inst 2017;36(2):2735e42. 

[15] Teng HH, Jiang ZL. On the transition pattern of the oblique

[37] Teng HH, Zhao W, Jiang ZL. A novel oblique detonation
structure and its stability. Chin Phys Letter 2007;

[38] Teng H, Zhang Y, Jiang Z. Numerical investigation on the induction zone structure of the oblique detonation waves. Computer Fluids 2014;95:127e31.

25] Gamezo VN, Khokhlov AM, Oran ES. The inﬂuence of shock bifurcations on shock-ﬂame interactions and DDT. Combust Flame 2001;126(4):1810e26.

26(2):3023e31.

30] Fusina G. Numerical investigation of oblique detonation

42. Teng, H. H., Y. N. Zhang, and Z. L. Jiang. 2014. Numerical investigation on the induction zone the oblique detonation waves. Comput. Fluids 95:127–31. doi:10.1016/j.
43–49. doi:10.1016/j.proci.2010.06.066.

Maeda, S., J. Kasahara, and A. Matsuo. 2012. Oblique detonation wave stability around a spherical projectile by a high time resolution optical observation. Combust. Flame 159 (2):887–96.

Maeda, S., S. Sumiya, J. Kasahara, and A. Matsuo. 2013. Initiation and sustaining mechanisms of Inst. 34:1973–80.
stabilized oblique detonation waves around projectiles. Proc. Combust. doi:10.1016/j.proci.2012.05.035.

Miao, S., J. Zhou, S. Liu, and X. Cai. 2018. Formation mechanisms and characteristics of transition patterns in oblique detonations. Acta Astronaut 142:121–29. doi:10.1016/j.actaastro.2017.10.035. 

38. Papalexandris, M. V. 2000. A numerical study of wedge-induced detonations. Combust. Flame 120

(4):526–38]

Silva, L. F. F. D., and B. Deshaies. 2000. Stabilization of an oblique detonation wave by a wedge:
a parametric numerical study. Combust. Flame 121 (1–2):152–66.
[Spalart, P., and S. Allmaras. 1992. A one-equation turbulence model for aerodynamic ﬂows. AIAA Paper 92–04
}
\end{tcolorbox}
\pagebreak

\bibliographystyle{elsarticle-num} 
\bibliography{MAIN_elsarticle}


\end{document}